\documentclass[letterpaper]{article} 
\usepackage{iclr2027_conference, times}
\usepackage[hyphens]{url}  
\usepackage{tikz}
\usepackage{xcolor}
\usepackage{pgfplots}
\pgfplotsset{compat=1.18}
\usepgfplotslibrary{groupplots}
\usetikzlibrary{positioning,arrows.meta,fit,backgrounds,calc,shapes.geometric,decorations.pathreplacing}
\usepackage{graphicx} 
\usepackage{natbib}  
\usepackage{caption} 
\usepackage{algorithm}
\usepackage{algorithmic}
\usepackage{amsmath, amssymb, bm}
\usepackage{multirow}

\usepackage{amsmath,amssymb,amsfonts,amsthm}
\usepackage{xcolor}

\def\our{GeoGAE}
\def\ours{GeoGAE}
\def\encod{\phi}
\def\decod{\psi}

\def\Beq#1\Eeq{\begin{equation}#1\end{equation}}
\def\Beqo#1\Eeqo{\begin{equation*}#1\end{equation*}}
\def\Beqs#1\Eeqs{\begin{align}#1\end{align}}
\def\Beqso#1\Eeqso{\begin{align*}#1\end{align*}}

\newtheorem{proposition}{Proposition}

\def\real{\mathbb R}
\def\natural{\mathbb N}
\def\setA{\mathbb A}

\def\comment#1{}

\newcommand{\res}[2]{$#1$ & ${}\pm{}$ & $#2$}
\newcommand{\best}[2]{$\mathbf{#1}$ & ${}\boldsymbol{\pm}{}$ & $\mathbf{#2}$}
\newcommand{\oom}{\multicolumn{3}{c|}{OOM}}
\newcommand{\na}{& \makebox[0pt]{N/A} &}

\usepackage{newfloat}
\usepackage{listings}
\DeclareCaptionStyle{ruled}{labelfont=normalfont,labelsep=colon,strut=off} 
\floatstyle{ruled}
\newfloat{listing}{tb}{lst}{}
\floatname{listing}{Listing}

\usepackage{booktabs}

\usepackage{wrapfig} 

\title{\our{}: Scalable Graph-Level Autoencoding via Hyperball Cloud Representations}
 \author{
     \!\!Radosław Nowak \\ 
     Institute of Theoretical and Applied Informatics \\ 
     Polish Academy of Sciences
     \And
     Anna Bielawska \\ 
     IDEAS Research Institute 
     \And
     Bogusz Stefańczyk \\ 
     IDEAS Research Institute 
     \And 
     Maciej Sanocki \\ 
     Faculty of Mathematics, Informatics and Mechanics \\
     Warsaw University of Technology 
     \And
     Paweł Wawrzyński \\ 
     IDEAS Research Institute
}

\iclrfinalcopy 
\def\arxivversion{{}}
\newcommand\figdir[1]{#1}

\begin{document}

\maketitle
\ifdefined\arxivversion
    \fancyhead[L]{}
\fi

\begin{abstract}
Embedding structured objects into Euclidean spaces has enabled a wide range of successful machine learning applications. Such objects include words, documents, image patches, time series, and graph nodes. In contrast, embedding entire graphs remains a challenging problem. Existing methods either sustain the original order of the graph nodes or match the output nodes to the input ones, both of which create scalability issues. In this work, we propose a~graph representation as a~cloud of hyperballs, which allows us to define a~specific---typically unique---node ordering. Based on this representation, we propose \our{}, an~autoencoder, in which the Transformer encoder translates a hyperball cloud into a~graph-level embedding, and the Transformer decoder translates the graph-level embedding back into the graph. This formulation enables the model to capture both the global graph structure and local relational patterns. We evaluate our method on multiple graph datasets, spanning various domains. The results demonstrate effectiveness of our method in encoding and reconstructing graphs from their embeddings.
\end{abstract}

\section{Introduction}

Fixed-size embeddings have proven very useful in a number of domains. Token embeddings are invaluable for natural language processing \citep{2024su+5}. Image embeddings are indispensable for image generation \citep{2022ramesh+4}. Time series embeddings are a useful tool in prediction \citep{2024foumani+4}. Embeddings are especially useful if any object in its domain can be reconstructed from its fixed-size vector representation. Then, the generation, transformation and optimization of these objects boil down to an equivalent operation in~$\mathbb{R}^m$, for a fixed $m$.  

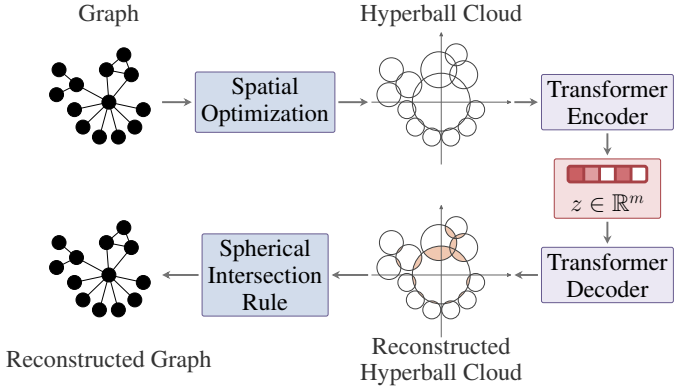
\begin{figure}[b!]
    \centering
    \vspace{-0.9em}
    \resizebox{9cm}{!}{
        \usetikzlibrary{positioning,arrows.meta,fit,backgrounds,calc}
    \definecolor{cInput}{HTML}{4C72B0}
    \definecolor{cBundle}{HTML}{55A868}
    \definecolor{cCls}{HTML}{DD8452}
    \definecolor{cTf}{HTML}{8172B3}
    \definecolor{cLatent}{HTML}{C44E52}
    
    \def\vO{0,0}             \def\rO{0.58}
    
    \def\vTOne{-0.05, 0.65}  \def\rTOne{0.36}
    \def\vTTwo{0.45, 0.55}   \def\rTTwo{0.28}
    \def\vTThree{0.3, 0.95}  \def\rTThree{0.22}
    
    \def\vLOne{-0.65, 0.35}  \def\rLOne{0.26}
    \def\vLTwo{-1.0, 0.65}   \def\rLTwo{0.24}
    \def\vLThree{-1.05, 0.15}\def\rLThree{0.24}
    
    \def\vPOne{-0.69, -0.15} \def\rPOne{0.18}
    \def\vPTwo{-0.52, -0.50} \def\rPTwo{0.18}
    \def\vPThree{-0.19, -0.70}\def\rPThree{0.18}
    \def\vPFour{0.19, -0.70} \def\rPFour{0.18}
    \def\vPFive{0.52, -0.50} \def\rPFive{0.18}
    \def\vPSix{0.69, -0.15}  \def\rPSix{0.18}

    \def\lens#1#2#3#4{\begin{scope}\clip (#1) circle (#2);
            \fill[cCls!45] (#3) circle (#4);\end{scope}}
    
    \def\ballclouduncolored#1{%
        \begin{scope}[shift={(#1)}]
            \begin{scope}[scale=1.5]
            \draw[->,black!60,line width=8pt] (-1.4,0) -- (1.4,0);
            \draw[->,black!60,line width=8pt] (0,-1.25) -- (0,1.4);
            
            \draw[draw=black!75,line width=6pt] (\vO) circle (\rO);
            \draw[draw=black!75,line width=6pt] (\vTOne) circle (\rTOne);
            \draw[draw=black!75,line width=6pt] (\vTTwo) circle (\rTTwo);
            \draw[draw=black!75,line width=6pt] (\vTThree) circle (\rTThree);
            \draw[draw=black!75,line width=6pt] (\vLOne) circle (\rLOne);
            \draw[draw=black!75,line width=6pt] (\vLTwo) circle (\rLTwo);
            \draw[draw=black!75,line width=6pt] (\vLThree) circle (\rLThree);
            \draw[draw=black!75,line width=6pt] (\vPOne) circle (\rPOne);
            \draw[draw=black!75,line width=6pt] (\vPTwo) circle (\rPTwo);
            \draw[draw=black!75,line width=6pt] (\vPThree) circle (\rPThree);
            \draw[draw=black!75,line width=6pt] (\vPFour) circle (\rPFour);
            \draw[draw=black!75,line width=6pt] (\vPFive) circle (\rPFive);
            \draw[draw=black!75,line width=6pt] (\vPSix) circle (\rPSix);
            \end{scope}
        \end{scope}%
    }

    \def\ballcloudcolored#1{%
        \begin{scope}[shift={(#1)}]
            \begin{scope}[scale=1.5]
            
            \lens{\vO}{\rO}{\vTOne}{\rTOne} 
            \lens{\vO}{\rO}{\vTTwo}{\rTTwo}
            \lens{\vO}{\rO}{\vLOne}{\rLOne}
            \lens{\vO}{\rO}{\vPOne}{\rPOne} 
            \lens{\vO}{\rO}{\vPTwo}{\rPTwo}
            \lens{\vO}{\rO}{\vPThree}{\rPThree} 
            \lens{\vO}{\rO}{\vPFour}{\rPFour}
            \lens{\vO}{\rO}{\vPFive}{\rPFive} 
            \lens{\vO}{\rO}{\vPSix}{\rPSix}
            
            \lens{\vTOne}{\rTOne}{\vTTwo}{\rTTwo} 
            \lens{\vTOne}{\rTOne}{\vTThree}{\rTThree} 
            \lens{\vTTwo}{\rTTwo}{\vTThree}{\rTThree}
            
            \lens{\vLOne}{\rLOne}{\vLTwo}{\rLTwo}
            \lens{\vLOne}{\rLOne}{\vLThree}{\rLThree}
            
            \draw[->,black!60,line width=8pt] (-1.4,0) -- (1.4,0);
            \draw[->,black!60,line width=8pt] (0,-1.25) -- (0,1.4);
            
            \draw[draw=black!75,line width=6pt] (\vO) circle (\rO);
            \draw[draw=black!75,line width=6pt] (\vTOne) circle (\rTOne);
            \draw[draw=black!75,line width=6pt] (\vTTwo) circle (\rTTwo);
            \draw[draw=black!75,line width=6pt] (\vTThree) circle (\rTThree);
            \draw[draw=black!75,line width=6pt] (\vLOne) circle (\rLOne);
            \draw[draw=black!75,line width=6pt] (\vLTwo) circle (\rLTwo);
            \draw[draw=black!75,line width=6pt] (\vLThree) circle (\rLThree);
            \draw[draw=black!75,line width=6pt] (\vPOne) circle (\rPOne);
            \draw[draw=black!75,line width=6pt] (\vPTwo) circle (\rPTwo);
            \draw[draw=black!75,line width=6pt] (\vPThree) circle (\rPThree);
            \draw[draw=black!75,line width=6pt] (\vPFour) circle (\rPFour);
            \draw[draw=black!75,line width=6pt] (\vPFive) circle (\rPFive);
            \draw[draw=black!75,line width=6pt] (\vPSix) circle (\rPSix);
            \end{scope}
        \end{scope}%
    }
    
    \def\minigraph#1{%
        \begin{scope}[shift={(#1)}]
            \begin{scope}[scale=1.5]
            \draw[gedge] (\vO)--(\vTOne); \draw[gedge] (\vO)--(\vTTwo);
            \draw[gedge] (\vO)--(\vLOne); 
            \draw[gedge] (\vO)--(\vPOne); \draw[gedge] (\vO)--(\vPTwo);
            \draw[gedge] (\vO)--(\vPThree); \draw[gedge] (\vO)--(\vPFour);
            \draw[gedge] (\vO)--(\vPFive); \draw[gedge] (\vO)--(\vPSix);
            
            \draw[gedge] (\vTOne)--(\vTTwo); 
            \draw[gedge] (\vTOne)--(\vTThree); 
            \draw[gedge] (\vTTwo)--(\vTThree);
            
            \draw[gedge] (\vLOne)--(\vLTwo);
            \draw[gedge] (\vLOne)--(\vLThree);
            
            \node[gdot] at (\vO){};
            \node[gdot] at (\vTOne){}; \node[gdot] at (\vTTwo){}; \node[gdot] at (\vTThree){};
            \node[gdot] at (\vLOne){}; \node[gdot] at (\vLTwo){}; \node[gdot] at (\vLThree){};
            \node[gdot] at (\vPOne){}; \node[gdot] at (\vPTwo){}; \node[gdot] at (\vPThree){};
            \node[gdot] at (\vPFour){}; \node[gdot] at (\vPFive){}; \node[gdot] at (\vPSix){};
            \end{scope}
        \end{scope}%
    }
    
    \begin{tikzpicture}[
        scale=12, transform shape,
        font=\huge,             
        >={Stealth[round, length=20mm, width=16mm]}, 
        gdot/.style   ={circle, fill=black, draw=black, line width=2pt, minimum size=2.8mm, inner sep=0pt}, 
        gedge/.style  ={draw=black, line width=10pt}, 
        proc/.style   ={rounded corners=20pt, draw=cTf!80!black, fill=cTf!16, line width=6pt,
            align=center, inner sep=5pt, minimum height=13mm, minimum width=32mm},
        proc2/.style  ={rounded corners=20pt, draw=cTf!80!black, fill=cInput!25, line width=6pt,
            align=center, inner sep=5pt, minimum height=13mm, minimum width=32mm}, 
        emb/.style    ={rounded corners=20pt, draw=cLatent!85!black, fill=cLatent!18,
            line width=6pt, align=center, inner sep=5pt, minimum height=13mm, minimum width=32mm},
        arr/.style    ={-{Stealth[length=30mm,width=30mm]}, line width=20pt, black!55},   
        lbl/.style    ={black!80},  
        ttl/.style    ={black!80},  
        ]
        
        \minigraph{(0,5.2)}
        \node[lbl] at (0,7.8) {Graph};
        
        \node[proc2] (opt) at (4.7,5.2) {Spatial\\ Optimization};
        
        \ballclouduncolored{(10.0,5.2)}
        \node[lbl] at (10.0,7.8) {Hyperball Cloud};
        
        \node[proc] (enc) at (15.0,5.2) {Transformer\\ Encoder};
        
        \draw[arr] (1.6,5.2) -- (opt.west);
        \draw[arr] (opt.east) -- (7.8,5.2);
        \draw[arr] (12.2,5.2) -- (enc.west);
        
        \node[emb] (gemb) at (15.0,2.6) {
            \begin{tikzpicture}[x=0.65em, y=0.65em, line width=0.12em, sharp corners]
                \fill[cLatent!90] (0,0) rectangle (1,1);
                \fill[cLatent!55] (1,0) rectangle (2,1);
                \fill[white]      (2,0) rectangle (3,1);
                \fill[cLatent!80] (3,0) rectangle (4,1);
                \fill[white]      (4,0) rectangle (5,1);
                
                \draw[cLatent!85!black, rounded corners=0.1em] (0,0) rectangle (5,1);
                
                \foreach \i in {1,2,3,4} {
                    \draw[cLatent!85!black] (\i,0) -- (\i,1);
                }
            \end{tikzpicture}\\[1pt]
            $z \in \mathbb{R}^m$
        };
        
        \draw[arr] (enc.south) -- (gemb.north);
        
        \node[proc] (dec) at (15.0,0) {Transformer\\ Decoder};
        \draw[arr] (gemb.south) -- (dec.north);
        
        \ballcloudcolored{(10.0,0)}
        \node[lbl, align=center] at (10.0,-2.6) {Reconstructed\\ Hyperball Cloud};
        
        \minigraph{(0,0)}
        \node[lbl] at (0,-2.6) {Reconstructed Graph};
        
        \draw[arr] (dec.west) -- (12.2,0);
        
        \node[proc2] (ssr) at (4.7,0) {Spherical\\ Intersection\\ Rule};
        \draw[arr] (7.8,0) -- (ssr.east);
        \draw[arr] (ssr.west) -- (1.6,0);
        
    \end{tikzpicture}
    }
    \caption{\textbf{Overview of GeoGAE.} The discrete input graph is first mapped into a~continuous hyperball cloud representation via spatial optimization and deterministic node ordering. A Transformer encoder then compresses this sequence into a fixed-size graph-level embedding $z~\in~\mathbb{R}^m$. In the decoding path, a Transformer decoder reconstructs the hyperball cloud from the latent vector, and the final discrete graph topology is exactly recovered using a static spherical intersection rule.}
    \label{fig:teaser}
\end{figure}

The field of graph neural networks (GNNs; \citealt{Corso2024GNNPrimer,Khemani2024GNNReview}) provides a~plethora of methods for conditional graph generation \citep{Ayadi2024UnifiedGuidance,Bian2024HierarchicalGraphLatentDiffusion,gao2025riemannian,Hou2024DAGVAE,ketata2025lift,li2025layerdag,liu2025beta,wang2025a,wen2025hyperplr,Wesego2025GraphDiffusion,You2024Latent3DGraphDiffusion,Zhou2024LatentGraphDiffusion}. However, the main purpose of these methods is typically to generate a diverse set of graphs possessing certain properties, rather than to reconstruct a specific, unique graph from a~given latent embedding. While some of these methods \citep{Wesego2025GraphDiffusion,You2024Latent3DGraphDiffusion,Zhou2024LatentGraphDiffusion} use node (and edge) embeddings as an auxiliary graph representation that could be padded to form a graph-level vector, this approach 
establishes a strict limit on the graph size.

Few works focus on a reversible, scalable transformation of fixed-size vectors into graphs. Methods proposed by \citet{2022malkowski+2} and \citet{bresson2026graviti} include the node order in the graph embedding. In contrast, \cite{winter2021permutation} and \citet{krzakala2025quest} ensure the graph embedding is independent of the node order, but at the cost of requiring complex matching mechanisms between the input and output graph nodes.

In this paper, we bridge this gap by designing a graph representation as a cloud of hyperballs in Euclidean space. This allows us to define a unique order of nodes. Our proposed graph-level autoencoder, intuitively described in Figure~\ref{fig:teaser}, uses the Transformer encoder to transform the sequence of nodes represented that way into a~fixed-size embedding, and the Transformer decoder to transform the embedding back to the sequence of hyperballs. The hyperballs are translated into the discrete graph structure with a~static rule. 
While representing nodes in 2- or 3-dimensional Euclidean space is a technique applied for molecular graph generation \citep{ketata2025lift}, we generalize the dimensionality to any~$d\in\natural$. 


\textbf{Contributions.} The main contributions of this work are threefold:
\begin{enumerate}
    \item \textbf{Continuous Hyperball Representation:} We introduce a continuous graph representation as hyperball clouds in Euclidean space, enabling canonical and deterministic node ordering.
    \item \textbf{Scalable Graph Autoencoder:} We propose \our{}, a Transformer-based autoencoder that compresses variable-sized hyperball clouds into fixed-size latent vectors $z \in \mathbb{R}^m$ and reconstructs original graph topologies using a static geometric rule.
    \item \textbf{Empirical Superiority:} We conduct extensive evaluations across diverse benchmark domains, demonstrating that \our{} achieves state-of-the-art topology reconstruction and scales to large graphs where existing baselines suffer from out-of-memory errors or severe~degradation.
\end{enumerate}

\section{Related work} 
\label{sec:related-work}

\paragraph{Graph to vector models.}
Graph-level representation learning aims to encode an entire graph into a fixed-dimensional vector for tasks such as classification or regression \citep{2024khoshraftar+1}. A~standard approach involves processing the graph through message-passing GNN layers \citep{2018velickovic+5,2019xu+3,2020chen+4} and applying a readout or pooling operation to aggregate node-level features into a global embedding. This includes simple pooling (e.g., sum, mean) \cite{2016kipf+1a}, attention-based pooling \citep{2019lee+2}, or hierarchical coarsening strategies \citep{2018ying+5,2019zhang+6,2019ma+3}. An alternative strategy introduces virtual nodes connected to all other nodes to directly capture global graph properties \citep{2017li+2,2019xu+3,2021brossard+2}. 

While these methods excel at predictive downstream tasks, their aggregation mechanisms inherently compress and entangle structural information, creating a severe information bottleneck. This irreversible loss of precise topological detail renders standard pooling and virtual node architectures poorly suited for exact graph reconstruction from the latent space. In contrast, our proposed \our{} framework replaces lossy aggregation with a reversible hyperball cloud representation, explicitly designed to preserve exact graph topology within a fixed-size vector.

\paragraph{Graph generation.} While the primary focus of \our{} is autoencoding, our decoder effectively functions as a~continuous-to-discrete graph generator. Most contemporary graph generation methods aim to learn the data distribution to sample novel structures, rather than exactly reconstructing specific inputs. Early approaches, such as GraphRNN \citep{2018you+4}, generated adjacency matrices sequentially using recurrent architectures. Recently, diffusion models have dominated the field \citep{liu2023generative_diffusion_graphs}, generating adjacency matrices similarly to images \citep{liu2025beta}, or utilizing autoregressive denoising steps \citep{Kong2023AutoregressiveDiffusionGraphs}. Several methods also generate node coordinates in Euclidean \citep{gao2025riemannian, You2024Latent3DGraphDiffusion} or non-Euclidean spaces \citep{fu2024HyperbolicGraphGen}, while others apply continuous-time diffusion over discrete states \citep{xu2024discrete_state_continuous_time_diffusion}. However, unlike our framework, which establishes a deterministic and reversible mapping between a unique graph and its latent embedding, these generative models primarily map prior noise distributions to graph distributions, making them unsuitable for exact topology preservation.

\paragraph{Graph autoencoders with embeddings in $\real^{n\times d}$.} These architectures implement the general structure of Variational Autoencoder (VAE: \citealt{2014kingma+1a}), with an~encoder that transforms the input graph into a~matrix of embeddings of its nodes. A~decoder performs the reverse transformation. Both encoder and decoder can take the form of message-passing GNN \citep{Imrie2020Deep3DLinker,Huang20223DLinker,Hou2024DAGVAE}. 




\paragraph{Graph autoencoders with embeddings in $\real^m$.} 
Our goal in this paper is to transform graphs of various sizes to embeddings of fixed size, $m$, and transform these embeddings back to graphs. A~simple way to achieve it is to assume that the number of vertices is bounded by a~fixed $n_{\max}$ and one of the autoencoders from the previous paragraph with an embedding of size $n_{\max}\times d$ with some kind of padding for smaller graphs. This idea is applied in GraphVAE \citep{2018simonovsky+1} for small graphs with $n_{\max}$ up to 38. 
\citet{2021hy+1} introduced MGVAE -- an autoencoder whose encoder recursively identifies clusters in the graph, and replace them with nodes of a~higher order graph. Eventually the input graph is reduced to a~fixed size embedding. The decoder recursively unpacks this embedding to the input graph. \citet{2022malkowski+2} presented ReGAE, an~autoencoder where the encoder recursively rolls up growing parts of the adjacency matrix into a~vector in $\real^m$, and the decoder, also recursively, unrolls the input matrix. While this architecture is able to reconstruct graphs with even thousands nodes, it is not node permutation equivariant which potentially leads to accuracy losses. In PIGVAE \citep{winter2021permutation} and GRALE \citep{krzakala2025quest}, the Transformer encoder produces permutation equivariant graph-level embedding and a~similar decoder reconstructs the graph. These designs raise the problem of assigning the output nodes to the input ones. PIGVAE leverages a learnable assigning modules fed with the graph embedding. GRALE extends this idea by combining an optimal-transport-based reconstruction loss with a Sinkhorn-based matcher which significantly improved stability and expressivity. GraViti \citep{bresson2026graviti} use a~similar Transformer based architecture and simply resigns from node ordering equivariance. In this paper, we dodge the problem of node order equivariance by introducing an~usually unique node order. Also, we use iteratively the transformer decoder to produce the output graph of an arbitrary size. 





\section{Method}




\subsection{Formal Problem Description} 
\label{sec:problem} 

We consider undirected, unweighted graphs without self-loops. Formally, a~graph is defined by a~set of vertices, $V$, and a~set of edges $E \subseteq \{\{v,u\}|v,u \in V, v\neq u\}$. We denote $n=|V|$ and optionally consider features of vertices and edges. Given a set $\mathcal{G}$ of such graphs and a target dimension $m \in \mathbb{N}$, our objective is to learn an $\text{encoder}: \mathcal{G} \to \mathbb{R}^m$ that compresses the graph into a fixed-size vector $z \in \mathbb{R}^m$, and a $\text{decoder}: \mathbb{R}^m \to \mathcal{G}$ that reconstructs this graph. Both mappings are end-to-end trained neural networks to minimize a~reconstruction error, generalizing effectively to unseen graphs.




\subsection{Model Overview}
\label{sec:overview}

Our proposed framework, \our{}, is illustrated in Figure~\ref{fig:teaser}. 
The pipeline first maps an input graph into a continuous hyperball cloud representation (details described in Section \ref{sec:ball-cloud}), where each hyperball corresponds to a graph node. Next, a Transformer encoder compresses this geometric sequence into a fixed-size latent vector $z \in \mathbb{R}^m$, constructed by concatenating $C$ learned class (\texttt{[CLS]}) tokens. Because these class tokens attend to the graph nodes without inter-\texttt{[CLS]} attention, each token is encouraged to specialize in capturing distinct topological properties.

In the decoding phase, a Transformer decoder interprets the latent embedding $z$ to sequentially reconstruct the hyperball cloud. Finally, the reconstructed cloud is mapped back into a discrete graph using the static spherical intersection rule (Section~\ref{sec:ball-cloud}). The entire architecture is trained end-to-end to encourage exact topological reconstruction.

\subsection{Hyperball cloud}
\label{sec:ball-cloud}




Let us consider representing graphs in a Euclidean space \citep{agrawal2021minimum,2024nowak+4}. In this formulation, graph nodes are mapped to spatial objects whose geometric distances reflect the underlying topological distances. The following proposition demonstrates that such a representation can be topologically lossless.

\begin{proposition}
\label{proposition1}
A graph with $n$ vertices can be losslessly represented as $n$ hyperballs of equal size in an $n^2$-dimensional space, such that two nodes are adjacent if and only if their corresponding hyperballs intersect.
\end{proposition}
\noindent The proof of Proposition~1 is provided in Appendix~A.

While lossless representation using equal-sized hyperballs is theoretically guaranteed, it requires an excessively high-dimensional space ($n^2$). This dimensionality bottleneck can be dramatically reduced by allowing hyperball radii to vary. Consider the hub topology in Figure~\ref{fig:teaser}, where a central node connects to 10 neighbors, with only one pair of neighbors being mutually adjacent. If all 11 nodes were represented by equal-sized hyperballs, packing 10 balls around the central one in low dimensions would geometrically force unwanted overlaps among the neighbors. Allowing variable radii bypasses this packing constraint, enabling exact, lossless representation even in low-dimensional space.

\paragraph{Translation of a Graph to a Hyperball Cloud.}
Graph nodes can be embedded in an Euclidean space as follows. Let $d\in\mathbb{N}$ be the dimension of this space, and $x^g_1, \dots, x^g_n\in\mathbb{R}^d$ be the geometric embeddings of the nodes. Let $x^g$ denote the matrix that gathers them. An embedding $x^g_i = [x^c_{i,1}, \dots, x^c_{i,d-1}, r_i]$ represents a hyperball in $(d-1)$-dimensional space with center $x^c_i$ and radius $f(r_i)$, where $f:\mathbb{R}\rightarrow\mathbb{R}_+$ is a strictly increasing function (e.g., Softplus).
Adjacent nodes correspond to intersecting balls, whereas non-adjacent nodes correspond to disjoint ones. The geometric embeddings are obtained by minimizing the loss function $\mathcal{L}_{\text{sphere}}$:
\Beqs   
    x^g & = \arg\min_{x^g} \mathcal{L}_{\text{sphere}}(x^g) 
    \label{opt:x} \\
    p_{i,j} & = \sigma\!\left( \frac{0.75\,(f(r_i)+f(r_j)) - \|x^c_i-x^c_j\|}{T} \right) 
    \label{opt:p} \\
    \mathcal{L}_{\text{sphere}}(x^g) & = \sum_{i,j} -\widehat\alpha_{i,j}\,(1-\widehat p_{i,j})^\gamma\log(\widehat p_{i,j}),
    \label{opt:L}
\Eeqs
where $\|\cdot\|$ denotes the standard $L_2$ Euclidean norm, $\sigma$ is the sigmoid function, $T > 0$ is a temperature scaling constant. The topological distance between nodes $i$ and $j$ is denoted by $D_{i,j}$. The target probability $\widehat p_{i,j}$ and class weight $\widehat\alpha_{i,j}$ are defined as $\widehat p_{i,j} = p_{i,j}$ and $\widehat\alpha_{i,j} = \alpha$ for adjacent nodes ($D_{i,j} = 1$), and $\widehat p_{i,j} = 1 - p_{i,j}$ and $\widehat\alpha_{i,j} = 1 - \alpha$ otherwise. The loss $\mathcal{L}_{\text{sphere}}$ treats hyperball intersection as a binary edge classification task. The class weight $\alpha$ mitigates class imbalance by prioritizing positive (adjacent) pairs, while the focusing parameter $\gamma$ down-weights well-classified node pairs to concentrate optimization on topologically ambiguous cases.

The optimization objective \eqref{opt:L} is invariant to rigid Euclidean transformations (translations and rotations); hence, its solution is not unique. To establish a canonical, permutation-invariant representation, we normalize the resulting geometric embeddings through the following steps:
\begin{enumerate}
\item We center the point cloud by shifting the coordinates such that the mean of all hyperball centers is zero ($\sum_i x^c_i = 0$).
\item We perform Singular Value Decomposition (SVD) on the centered spatial coordinates to find the principal axes, and re-express the embeddings within this new coordinate system.
\item To resolve reflectional ambiguities along the principal axes, we evaluate the third moment (skewness) of the projected coordinates. Specifically, for any axis $k \in \{1,\dots,d-1\}$, if $\sum_i (x^c_{i,k})^3 < 0$, we negate the $k$-th coordinate across all node embeddings.
\end{enumerate}
Barring highly symmetrical graph topologies, this deterministic normalization yields a unique set of geometric embeddings for any given graph. In our experiments we choose the smallest geometric embedding size which scores 100\% F1 on all dataset graphs.


\textbf{Translation of Hyperball Cloud to Graph.} Geometric embeddings that minimize \eqref{opt:L} enable determining if there is an edge between the $i$-th and $j$-th node using a direct intersection rule. Given the hyperball centers $x_i, x_j$ and their raw radii $r_i, r_j$, the margin is calculated as the difference between the scaled sum of the radii and the Euclidean distance between the centers:
$$m_{i,j} = 0.75 (f(r_i) + f(r_j)) - \|x_i - x_j\|_2$$
where $f$ denotes the softplus function. The decoding rule is then strictly defined by the sign of the margin:
\begin{equation}
    \text{if } m_{i,j} > 0, \text{ then there is an edge; else, not.}
\end{equation}
During training, the margin $m_{i,j}$ is scaled and forms logits for the binary cross-entropy and focal loss terms. At inference, the rule explicitly relies on the geometric intersection ($m_{i,j} > 0$), eliminating the need to optimize an arbitrary threshold on the training set.

\subsection{Node ordering and bundles}
\label{sec:node-ordering}

\paragraph{Node ordering.} The node ordering procedure begins by computing the geometric center (mean) of all hyperball centers. The first node in the sequence is chosen as the one whose hyperball center is closest to this global center. The consecutive nodes are then selected greedily by minimizing a distance to the previously selected node, and an additional penalty. Specifically, the Euclidean distance between ball centers is scaled by the average of their radii, and incremented by a scaled base-2 logarithm of the candidate node's distance to the global center. This ensures that nodes are ordered primarily by their relative spatial proximity and sizes, with a slight preference for nodes closer to the center of the entire structure. Algorithm 1 in Appendix~B presents the details of this procedure. Its result is denoted by $(y^g_1, \dots, y^g_n)$.



\paragraph{Bundles.} 
\label{sec:bundles}
To ensure scalability and mitigate the quadratic computational complexity inherent to the self-attention mechanism in Transformers, consecutive nodes from the ordered sequence are grouped into contiguous blocks called \textit{bundles} of size $b \in \mathbb{N}$. When the total number of nodes $N$ is not evenly divisible by $b$, the remaining slots in the final bundle are padded by repeating the geometric embedding of the last node. Each bundle of $b$ geometric embeddings in $\mathbb{R}^{b \times d}$ is flattened and projected into a single $d_{\text{tok}}$-dimensional sequence token using a trainable Multilayer Perceptron (MLP) ($\text{Bundle}\to\text{Token}$). This enables nodes within a single bundle to be processed by the encoder and generated by the decoder simultaneously. For smaller datasets where scalability is not a bottleneck, we set $b=1$, which bypasses bundling and represents each node with an individual token. See Appendix~\ref{app:bundles} for exact tensor dimensions and decoder de-bundling details.

\subsection{Architecture}

\paragraph{Encoder.}

\begin{figure*}[t]
  \centering
  \resizebox{\textwidth}{!}{
        \definecolor{cInput}{HTML}{4C72B0}
\definecolor{cBundle}{HTML}{55A868}
\definecolor{cCls}{HTML}{DD8452}
\definecolor{cTf}{HTML}{8172B3}
\definecolor{cLatent}{HTML}{C44E52}
\definecolor{cAttn}{HTML}{4E79A7}
\definecolor{cFfn}{HTML}{59A14F}

\begin{tikzpicture}[
    font=\normalsize,
    >={Stealth[round]},
    box/.style       ={rounded corners=2.5pt, draw, semithick, align=center, inner sep=4pt},
    ball/.style      ={circle, draw=cInput!85!black, fill=cInput!25, semithick},
    mlp/.style       ={box, fill=cBundle!12, draw=cBundle!80!black, minimum width=24mm, minimum height=11mm},
    tok/.style       ={box, fill=cBundle!18, draw=cBundle!80!black, minimum width=12mm, minimum height=8mm},
    cls/.style       ={box, fill=cCls!28, draw=cCls!85!black, minimum width=12mm, minimum height=8mm},
    latent/.style    ={box, fill=cLatent!22, draw=cLatent!85!black, minimum width=12mm, minimum height=10mm},
    sub/.style       ={box, minimum height=12mm, inner sep=4pt},
    add/.style       ={circle, draw, semithick, inner sep=0pt, minimum size=5.5mm},
    arr/.style       ={->, semithick, black!65},
    thin arr/.style  ={->, black!45},
    res/.style       ={->, semithick, cLatent!70!black, dashed},
    capt/.style      ={font=\large, black!55},
    tag/.style       ={font=\large, black!60},
    ]

    \node[ball,minimum size=6.5mm] (b1) at (0,3.55) {};
    \node[ball,minimum size=5mm]   (b2) at (0,2.95) {};
    \node[ball,minimum size=7.5mm] (b3) at (0,2.20) {};
    \node[ball,minimum size=5mm]   (b4) at (0,1.60) {};

    \node[ball,minimum size=6mm]   (b5) at (0,0.40) {};
    \node[ball,minimum size=5mm]   (b6) at (0,-0.20) {};

    \node[capt, align=center] at (0,-1.6)
    {ordered geometric\\ embeddings $(y^g_1,\dots,y^g_n)$};
    \node[tag,rotate=90] at (-1.1,1.95) {center $\oplus$ radius};

    \node[mlp] (m1) at (2.1,3.25) {Bundle$\to$Token\\ MLP};
    \node[mlp] (m2) at (2.1,1.90) {Bundle$\to$Token\\ MLP};

    \node at (2.1,1.15) {$\vdots$};
    \node[mlp] (m3) at (2.1,0.10) {Bundle$\to$Token\\ MLP};

    \node[tag, align=center] at (2.1, 4.45) {shared weights,\\ $\lceil n{/}k\rceil$ bundles};

    \draw[thin arr] (b1) -- (m1); \draw[thin arr] (b2) -- (m1);
    \draw[thin arr] (b3) -- (m2); \draw[thin arr] (b4) -- (m2);
    \draw[thin arr] (b5) -- (m3); \draw[thin arr] (b6) -- (m3);

    \begin{scope}
        \foreach \dx/\dy in {0.20/0.20, 0.10/0.10}{
            \node[cls, text opacity=0] at (5.5+\dx,4.25+\dy) {[CLS]$_{1:C}$};
        }
    \end{scope}
    \node[cls] (cls) at (5.5,4.25) {[CLS]$_{1:C}$};

    \node[tag, text=cCls!85!black, inner sep=1pt, align=center] (learned) at (2.6, 5.7) {learned\\parameters};
    \draw[->, semithick, cCls!85!black, shorten >=2pt] (learned.east) -- (cls.north west);

    \node[tok] (t1)  at (5.5,3.25) {$t_1$};
    \node[tok] (t2)  at (5.5,1.90) {$t_2$};

    \node at (5.5,1.15) {$\vdots$};

    \node[tok] (t3)  at (5.5,0.10) {$t_T$};

    \draw[arr] (m1) -- (t1);
    \draw[arr] (m2) -- (t2);
    \draw[arr] (m3) -- (t3);

    \node[draw=black!35, rounded corners=3pt, dashed, inner sep=6pt,
    fit=(cls)(t1)(t3), label={[capt, yshift=2mm]above:input sequence}] (seq) {};

    \begin{scope}
        \foreach \dx/\dy in {0.28/0.28, 0.14/0.14, 0/0}{
            \node[box, fill=cTf!16, draw=cTf!80!black, minimum width=29mm,
            minimum height=44mm] at (9.0+\dx,2.35+\dy) {};
        }
    \end{scope}
    \node[align=center, text=cTf!30!black, font=\large] at (9.0,3.0)
    {Transformer\\ Encoder};

    \node[align=center, text=cTf!35!black] at (9.0,1.6) {$\times\,L$\\[3pt] pre-norm};

    \node[tag, align=center] at (9.0,-1.0)
    {node$\leftrightarrow$node full attention\\ (permutation-invariant);\\ CLS restricted};

    \draw[arr] (seq.east |- 0,2.35) -- (9.0-1.45,2.35);

    \node[sub, fill=black!5, draw=black!45, minimum width=18mm, minimum height=8mm] (lnf) at (13.0,2.35) {LayerNorm};
    
    \begin{scope}
        \foreach \dx/\dy in {0.20/0.20, 0.10/0.10}{
            \node[cls, text opacity=0] at (15.5+\dx,2.35+\dy) {[CLS]$'_{1:C}$};
        }
    \end{scope}
    \node[cls] (clso) at (15.5,2.35) {[CLS]$'_{1:C}$};
    \node[latent, font=\Large] (z)  at (15.5,1.00) {$z$};

    \draw[arr] (9.0+0.28+1.45,2.35) -- (lnf);
    \draw[arr] (lnf) -- (clso);
    \draw[arr] (clso) -- (z);

    \node[tag, align=center] at (15.5,-0.35)
    {latent code\\ $z=\mu+\sigma\odot\epsilon$ (VAE, opt.)};

    \coordinate (zanchor) at (1.5,-4.8);
    \node[capt, font=\Large] (zin) at (zanchor) {$x$};

    \node[sub, fill=black!5,  draw=black!45, minimum width=16mm,
    right=4mm of zin]  (zln1) {LayerNorm};

    \node[sub, fill=cAttn!16, draw=cAttn!80!black, text width=34mm,
    right=4mm of zln1] (zatt) {Multi-Head Self-Attn\\ RoPE on $Q,K$};

    \node[add, right=4mm of zatt] (za1) {$+$};

    \node[sub, fill=black!5,  draw=black!45, minimum width=16mm,
    right=4mm of za1]  (zln2) {LayerNorm};

    \node[sub, fill=cFfn!16,  draw=cFfn!80!black, text width=34mm,
    right=4mm of zln2] (zff)  {Feed-Forward\\ Linear - GELU - Linear};

    \node[add, right=4mm of zff] (za2) {$+$};
    \node[capt, right=4mm of za2, font=\Large] (zout) {$x'$};

    \draw[arr] (zin)  -- (zln1);
    \draw[arr] (zln1) -- (zatt);
    \draw[arr] (zatt) -- (za1);
    \draw[arr] (za1)  -- (zln2);
    \draw[arr] (zln2) -- (zff);
    \draw[arr] (zff)  -- (za2);
    \draw[arr] (za2)  -- (zout);

    \coordinate (r1) at ($(zin.east)!0.5!(zln1.west)$);
    \coordinate (r2) at ($(za1.east)!0.5!(zln2.west)$);
    \coordinate (rapex) at ($(zanchor)+(0,1.6)$);
    \draw[res] (r1) -- ++(0,1.6) -| (za1.north);
    \draw[res] (r2) -- ++(0,1.6) -| (za2.north);

    \node[draw=cTf!55, rounded corners=4pt, dotted, semithick, inner sep=8pt,
    fit=(zin)(zln1)(zatt)(za1)(zln2)(zff)(za2)(zout)(rapex),
    label={[tag,text=cTf!35!black, yshift=0mm, font=\Large]above:encoder layer}] (zoombox) {};

    \draw[cTf!55, dotted, semithick] (9.0-1.4,0.15) -- (zoombox.north west);
    \draw[cTf!55, dotted, semithick] (9.0+1.4,0.15) -- (zoombox.north east);

\end{tikzpicture}
    }
  \caption{Geometric graph transformer encoder. The input graph is represented as an ordered sequence of hyperballs and mapped into sequence tokens. A full, permutation-invariant self-attention mechanism processes these tokens alongside $C$ dedicated \texttt{[CLS]} tokens. The output representations of the \texttt{[CLS]} tokens are concatenated to yield the global graph embedding $z \in \mathbb{R}^m$.}
  \label{fig:encoder}
\end{figure*}
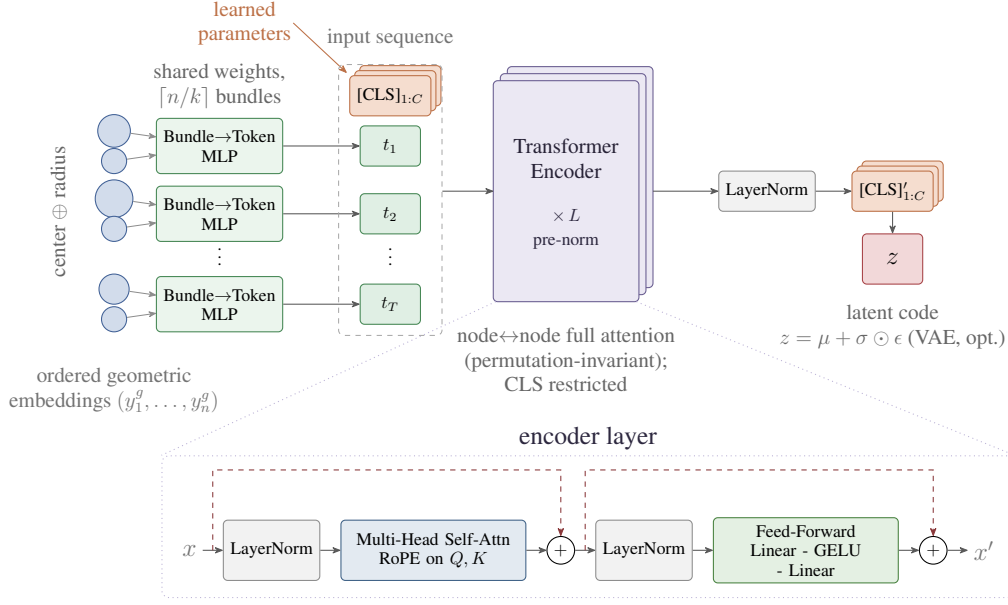

The encoder of \ours{} (Figure~\ref{fig:encoder}) compresses the input graph into a fixed-size latent representation through a streamlined pipeline. First, the graph is mapped into a hyperball cloud, deterministically ordered, and grouped into $T = \lceil n/b \rceil$ contiguous bundles of size $b$ (Sections~\ref{sec:ball-cloud}--\ref{sec:bundles}). An MLP, denoted $\encod$, maps each bundle into a sequence token (details provided in Appendix \ref{app:bundles}). To preserve relative positional information without explicit absolute position embeddings, Rotary Positional Embeddings (RoPE) are applied directly to queries and keys within each self-attention layer. The sequence is prepended with $C$ trainable \texttt{[CLS]} tokens and processed by the Transformer encoder (attention masking details are provided in Appendix~\ref{app:encoder_masking}). Finally, the output representations of the $C$ \texttt{[CLS]} tokens are extracted and concatenated to form the global graph embedding $z \in \mathbb{R}^m$, optionally followed by a~VAE reparameterization.

\begin{figure}[t]
  \centering
  \resizebox{\textwidth}{!}{
\usetikzlibrary{positioning,arrows.meta,fit,backgrounds,calc,shapes.geometric}
\definecolor{cInput}{HTML}{4C72B0}
\definecolor{cBundle}{HTML}{55A868}
\definecolor{cCls}{HTML}{DD8452}
\definecolor{cTf}{HTML}{8172B3}
\definecolor{cLatent}{HTML}{C44E52}
\definecolor{cAttn}{HTML}{4E79A7}
\definecolor{cFfn}{HTML}{59A14F}

\def\ballcloud#1{%
    \begin{scope}[shift={(#1)}]
        \draw[->,black!40,thin] (-1.15,0) -- (1.15,0);
        \draw[->,black!40,thin] (0,-1.15) -- (0,1.15);
        \draw[draw=black!75,semithick] (0,0.05)     circle (0.46);
        \draw[draw=black!75,semithick] (0.08,0.60)  circle (0.27);
        \draw[draw=black!75,semithick] (0.50,0.40)  circle (0.27);
        \draw[draw=black!75,semithick] (0.78,0.06)  circle (0.23);
        \draw[draw=black!75,semithick] (0.02,-0.59) circle (0.26);
        \draw[draw=black!75,semithick] (-0.56,-0.18)circle (0.27);
        \draw[draw=black!75,semithick] (-0.58,0.22) circle (0.27);
    \end{scope}%
}
\begin{tikzpicture}[
    font=\large,
    >={Stealth[round]},
    box/.style   ={rounded corners=2.5pt, draw, semithick, align=center, inner sep=4pt},
    mlp/.style   ={box, fill=cBundle!12, draw=cBundle!80!black, minimum width=24mm, minimum height=11mm},
    tok/.style   ={box, fill=cBundle!18, draw=cBundle!80!black, minimum width=12mm, minimum height=8mm},
    cls/.style   ={box, fill=cCls!28, draw=cCls!85!black, minimum width=12mm, minimum height=8mm},
    latent/.style ={box, fill=cLatent!22, draw=cLatent!85!black, minimum width=13mm, minimum height=11mm},
    sub/.style   ={box, minimum height=9mm, inner sep=3pt},
    add/.style   ={circle, draw, semithick, inner sep=0pt, minimum size=5.5mm},
    arr/.style   ={->, semithick, black!65},
    memarr/.style ={->, semithick, cLatent!70!black, dashed},
    res/.style   ={->, semithick, cLatent!70!black, dashed},
    capt/.style  ={font=\Large, black!55}, 
    tag/.style   ={font=\Large, black!60}, 
    ]
    
    \node[latent, font=\LARGE] (z) at (-0.15,2.15) {$z$};
    \node[capt,text width=50mm,align=center] at (-0.15,3.40)
    {graph embedding\\ (latent memory)};
    
    \node[sub, fill=black!5, draw=black!45, text width=30mm, align=center, font=\large] (q) at (-0.15,-0.35)
    {learned queries\\ / start token};
    
    \begin{scope}
        \foreach \dx/\dy in {0.28/0.28, 0.14/0.14, 0/0}{
            \node[box, fill=cTf!16, draw=cTf!80!black, minimum width=26mm,
            minimum height=38mm] at (3.6+\dx,0.76+\dy) {};
        }
    \end{scope}
    \node[align=center, text=cTf!30!black, font=\large] at (3.6,1.31)
    {Transformer\\ Decoder};
    \node[align=center, text=cTf!35!black, font=\large] at (3.6,0.16) {$\times\,M$\\[1pt]
        {\normalsize pre-norm}};
    
    \draw[memarr] (z.east) -- (3.6-1.32,2.15);
    \draw[arr]    (q.east) -- (3.6-1.32,-0.35) node[midway,below,tag]{tgt};
    
    \node[tok] (t1) at (6.2,2.4) {$t_1$};
    \node[tok] (t2) at (6.2,0.90) {$t_2$};
    \node at (6.2,0.00) {$\vdots$}; 
    \node[tok] (t3) at (6.2,-0.95) {$t_T$}; 
    
    \draw[arr] (3.6+0.28+1.3,0.90) -- (5.55,0.90);
    
    \node[draw=black!35, rounded corners=3pt, dashed, inner sep=6pt,
    fit=(t1)(t3), label={[capt, yshift=1mm]above:output sequence}] (seq) {};
    
    \node[mlp] (m1) at (10.0,2.4) {Token$\to$Nodes\\ MLP};
    \node[mlp] (m2) at (10.0,0.90) {Token$\to$Nodes\\ MLP};
    \node at (10.0, 0.00) {$\vdots$}; 
    \node[mlp] (m3) at (10.0,-0.95) {Token$\to$Nodes\\ MLP}; 
    
    \node[tag, text width=62mm, align=center] at (10.0,3.7) {shared, $k$ hyperballs \\ per~token};
    \draw[arr] (t1) -- (m1);
    \draw[arr] (t2) -- (m2);
    \draw[arr] (t3) -- (m3);
    
    \ballcloud{(14.0,0.90)}
    \node[capt,text width=42mm,align=center] at (14,-1.15)
    {hyperball cloud\\ $(\hat y^g_1,\dots,\hat y^g_{\hat n})$, center\,$\oplus$\,radius};
    
    \draw[arr, shorten >= 4pt] (m1.east) -- (14.0-1.15,1.50);
    \draw[arr, shorten >= 4pt] (m2.east) -- (14.0-1.15,0.90);
    \draw[arr, shorten >= 4pt] (m3.east) -- (14.0-1.15,0.2); 
    
    \node[sub, fill=black!5, draw=black!45, text width=24mm, align=center, font=\large]
    (stop) at (6.2,-2.45) {stop head}; 
    \draw[arr] (t3.south) -- (stop.north);
    \draw[arr] (stop.east) -- ++(2.2,0)
    node[midway,above,tag]{$\hat n$} node[right,tag]{graph size};
    
    \draw[arr] (14.2+1.15,0.90) -- ++(1.1,0)
    node[right,tag, align=center]{static spherical\\ rule $\Rightarrow$ graph};
    
    \coordinate (zanchor) at (1.5,-5.5); 
    \node[capt, font=\LARGE] (zin) at (zanchor) {$x$};
    \node[sub, fill=black!5, draw=black!45, minimum width=13mm,
    right=3mm of zin]  (zln1) {LayerNorm};
    \node[sub, fill=cAttn!16, draw=cAttn!80!black, text width=18mm,
    right=3mm of zln1] (zsa)  {Masked Self-Attn\\ {\normalsize RoPE}};
    \node[add, right=3mm of zsa] (za1) {$+$};
    \node[sub, fill=black!5, draw=black!45, minimum width=13mm,
    right=3mm of za1]  (zln2) {LayerNorm};
    \node[sub, fill=cCls!18, draw=cCls!80!black, text width=25mm,
    right=3mm of zln2] (zca)  {Cross-Attn\\ {\normalsize memory \large $z$, \normalsize no PE}};
    \node[add, right=3mm of zca] (za2) {$+$};
    \node[sub, fill=black!5, draw=black!45, minimum width=13mm,
    right=3mm of za2]  (zln3) {LayerNorm};
    \node[sub, fill=cFfn!16, draw=cFfn!80!black, text width=18mm,
    right=3mm of zln3] (zff)  {Feed-Forward\\ {\normalsize GELU}};
    \node[add, right=3mm of zff] (za3) {$+$};
    \node[capt, right=3mm of za3, font=\LARGE] (zout) {$x'$};
    
    \draw[arr] (zin)--(zln1); \draw[arr] (zln1)--(zsa); \draw[arr] (zsa)--(za1);
    \draw[arr] (za1)--(zln2); \draw[arr] (zln2)--(zca); \draw[arr] (zca)--(za2);
    \draw[arr] (za2)--(zln3); \draw[arr] (zln3)--(zff); \draw[arr] (zff)--(za3);
    \draw[arr] (za3)--(zout);
    
    \node[latent, font=\LARGE] (zmem) at ($(zca)+(0,-1.9)$) {$z$};
    \draw[memarr] (zmem) -- (zca);
    
    \coordinate (r1) at ($(zin.east)!0.5!(zln1.west)$);
    \coordinate (r2) at ($(za1.east)!0.5!(zln2.west)$);
    \coordinate (r3) at ($(za2.east)!0.5!(zln3.west)$);
    \coordinate (rapex) at ($(zanchor)+(0,1.0)$);
    \draw[res] (r1) -- ++(0,1.0) -| (za1.north);
    \draw[res] (r2) -- ++(0,1.0) -| (za2.north);
    \draw[res] (r3) -- ++(0,1.0) -| (za3.north);
    
    \node[draw=cTf!55, rounded corners=4pt, dotted, semithick, inner sep=8pt,
    fit=(zin)(zln1)(zsa)(za1)(zln2)(zca)(za2)(zln3)(zff)(za3)(zout)(rapex)(zmem),
    label={[tag,text=cTf!35!black, font=\LARGE]above left:decoder layer}] (zoombox) {};
    
    \draw[cTf!55, dotted, semithick] (3.6-1.3,-1.14) -- (zoombox.north west);
    \draw[cTf!55, dotted, semithick] (3.6+1.3,-1.14) -- (zoombox.north east);
    
\end{tikzpicture}
    }
  \caption{Geometric graph transformer decoder. Acts as a generative module that reconstructs the graph. It uses the latent vector $z$ as context (latent memory) within its \textit{cross-attention} layers, combining it with learned queries. The decoder generates a sequence of output tokens. Next, Multilayer Perceptron heads (Token$\to$Nodes MLP) map these tokens back into a hyperball cloud. Concurrently, a dedicated stop head predicts the target graph size, and a geometric mechanism (static spherical rule) reconstructs the final edges between the nodes.}
  \label{fig:decoder}
\end{figure}
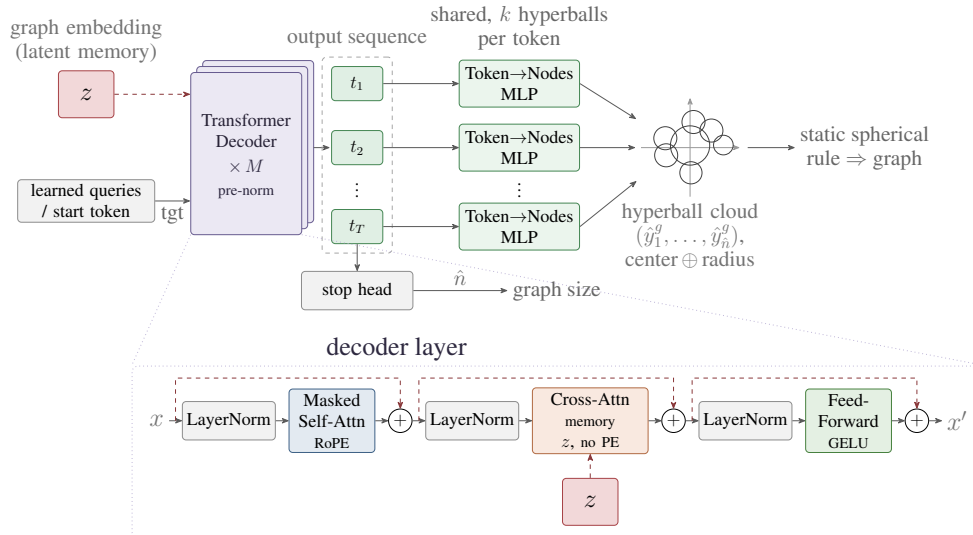

\paragraph{Decoder.}\label{sec:transformer+head}The Transformer decoder (Figure~\ref{fig:decoder}) reconstructs the hyperball cloud by conditioning on the latent vector $z$ via cross-attention over sequence queries. Instead of a standard linear-softmax head, a prediction module $\decod$ (detailed in Appendix~\ref{app:bundles}) processes each decoder output token to yield two distinct outputs: (i)~a matrix of $b$ geometric node embeddings containing their spatial centers and radii, and (ii)~$b$ independent binary stopping logits. Each logit corresponds to an individual node within the bundle and predicts whether it terminates the graph. This per-node stopping mechanism provides exact node-level precision, allowing the model to halt generation and cleanly truncate the final bundle when the graph size is not a multiple of $b$.





\subsection{Multi-layer teacher forcing} 
\label{sec:ML-TF}

Transformer decoders are typically trained via teacher forcing, utilizing the right-shifted ground truth sequence---achieved by prepending a~special `start' vector---to predict the full target sequence terminating with an~`end' vector. However, this paradigm is notoriously susceptible to exposure bias; the absence of ground truth data during inference causes cascading errors when the model is conditioned on its own flawed predictions.

To counteract this unwanted byproduct, we propose an iterative refinement strategy by stacking the decoder $K \in \mathbb{N}$ times (with weight sharing of the instances). The base instance is conditioned on the right-shifted target sequence, while each successive instance takes the output sequence of the underlying layer, similarly right-shifted by prepending the `start' vector, as its input. All $K$ instances are jointly supervised to output the complete sequence ending with the `end' vector, effectively bridging the gap between the training and inference regimes.

\subsection{Node and edge attributes} 

When present, node attributes are concatenated into each node's geometric embedding before bundling -- this effectively widens the input that $\encod$ consumes. On the other side, the decoder produces output tokens, which are then consumed by a separate MLP trained to reconstruct each node's attributes. 

Edge attributes are consumed directly by the encoder's self-attention, in every layer, rather than as a per-token input channel. When a pair of tokens is connected via an edge, the edge attribute is concatenated into the token's key, and then consumed by a dedicated per-layer linear projection that is taught to correct the key, and the value only for those pairs. In the inference mode, the decoder, after deconstructing the adjacency matrix of the output graph uses a separate MLP, which for each connected pairs of tokens, takes a vector of their sum and absolute difference as an input, and produces a vector of their edge attribute. During the training process, the decoder uses the ground truth graph edges.


{\bf The total loss} minimized in the training of our proposed architecture is specified in Appendix~B. 

\section{Experimental study}

We evaluate \ours{} across seven benchmark datasets spanning chemical, social, and synthetic domains. Our experiments assess topological and feature reconstruction fidelity, scalability on large graphs, and the impact of key architectural choices via ablations.

\paragraph{Benchmarks.}


\begin{table}
\centering
\begin{tabular}{l|r|r|r}
Dataset & Graphs & Avg. nodes & Avg. edges \\ 
\hline
MUTAG & 188 & 17.90 & 19.80 \\ 
AIDS & 2,000 & 15.69 & 16.20 \\ 
IMDB-BIN & 1,000 & 19.77 & 96.53 \\ 
QM9 & 129,433 & 18.03 & 18.63 \\ 
SYNTHETIC NEW & 300 & 100.00 & 196.25 \\ 
COLLAB & 5,000 & 74.49 & 2,457.78 \\ 
REDDIT-BIN & 2,000 & 429.63 & 497.75 \\ 
\end{tabular}
\caption{Basic statistics of datasets used in our benchmark.}
\label{tab:datasets}
\end{table}

We use seven datasets of graphs of various nature, including social, chemical and synthetic ones. Along with their basic statistics, they are listed in Table~\ref{tab:datasets} and characterized more broadly in Appendix~C. 
These datasets provide the raw graph structures, which we process to benchmark our model's ability to generate graph-level embeddings and reconstruct the input graphs. We divide these datasets into training, validation, and test subsets in a~70:15:15 ratio.

\paragraph{Baselines.} 

Among the autoencoders capable of generalizing to unseen graphs---meaning they can be trained on one dataset and evaluated on another---are PIGVAE \citep{winter2021permutation}, ReGAE \citep{2022malkowski+2}, and GRALE \citep{krzakala2025quest}. Our experiments aim to train these models to encode and reconstruct graph topologies using a~training subset, verifying their performance on a~held-out test subset.

\subsection{Training}

For each dataset, graph embedding dimensions were scaled proportionally with average graph complexity, with \ours{} latent sizes aligned directly with ReGAE to ensure a fair comparison on high-capacity architectures (Table~\ref{tab:our:hyperparameters}). Hyperparameters for all baselines were systematically optimized via dataset-specific random search, as detailed in Tables~\ref{tab:our:hyperparameters}, \ref{tab:pigvae:hyperparameters}, \ref{tab:regae:hyperparameters}, and~\ref{tab:grale:hyperparameters}. For PIGVAE and GRALE, embedding sizes were explicitly included in hyperparameter tuning; larger latent dimensions degraded performance or induced training instability. Throughout all experiments, the temperature parameter in Eq.~\ref{opt:p} was fixed at $T = 0.4$.

To guarantee statistical reliability, all models were trained and evaluated across five independent runs using distinct fixed seeds and pre-computed data splits. We report the mean and standard deviation across these runs.

\begin{table*}
\centering
\begin{tabular}{l|r|r|r|r|r|r|r|r|r}
Dataset & \!tok-d\! & emb & \!geom-d\! & encod\;\; & decod\;\; & lr\;\; & \!enc-h\! & \!dec-h\! & batch\\
\hline
MUTAG        & 64 & 128 & 4 & $1024 \times 4$ & $256 \times 2$ & 1e-3 & 8 &  1 &  2 \\
AIDS         & 64 & 192 & 6 & $1024 \times 4$ & $512 \times 2$ & 1e-3 & 8 & 1 & 32 \\
IMDB-BIN     & 64 & 192 & 6 & $512 \times 8$ & $1024 \times 6$ & 1e-3 & 2 & 1 & 32 \\
QM9          & 64 & 192 & 4 & $1024 \times 6$ & $1024 \times 4$ & 1e-3 & 4 & 4 & 256 \\
SYNTH. NEW    & 64 & 192 & 6 & $2048 \times 8$ & $1024 \times 2$ & 5e-4 & 8 & 4 & 8 \\
COLLAB       & 96 & 576 & 9 & $256 \times 8$ & $2048 \times 4$ & 1e-3 & 4 &  1 & 32 \\ 
REDDIT-BIN   & 144 & 1728 & 9 & $512 \times 4$ & $1024 \times 4$ & 1e-3 & 6 & 4 & 16
\end{tabular}
\caption{Hyperparameters of \our{}: tok-d -- token dimension, emb -- graph embedding size (integer multiple of tok-d), geom-d -- dimension of the geometric embedding, encod -- width and depth (width $\times$ number of layers) of the encoder hidden layers, decod -- width and depth (width $\times$ number of layers) of the decoder hidden layers, lr -- learning rate, enc-h -- number of attention heads in the encoder, dec-h -- number of attention heads in the decoder, batch -- batch size. All datasets except REDDIT-BINARY use a bundle size of 1; REDDIT-BINARY uses a bundle size of 4. 
}
\label{tab:our:hyperparameters}
\vspace{-1.1em}
\end{table*}

\subsection{Results}

Table~\ref{tab:test_f1} summarizes the edge reconstruction F1 scores, weighted by vertex count across test graphs. \ours{} consistently matches or outperforms baseline architectures on smaller graphs while demonstrating superior scalability on larger benchmarks. On complex, high-density datasets such as REDDIT-BINARY and COLLAB, \ours{} achieves $56.5\%$ and $90.9\%$ F1, respectively, whereas PIGVAE, ReGAE, and GRALE suffer from training instability or degenerate into trivial predictions (e.g., predicting all $0$s or $1$s).

Table~\ref{tab:test_f1_features} reports joint topology and feature reconstruction performance, using F1 for discrete attributes and RMSE for continuous properties. While baselines suffer severe trade-offs, such as GRALE's high feature error on QM9 ($5.4$ node RMSE) or PIGVAE's topological collapse ($25.6\%$ F1), \ours{} maintains a robust balance. On AIDS, \ours{} achieves superior feature fidelity across both categorical ($85.0\%$ edge F1) and continuous attributes ($0.8$ node RMSE) without sacrificing topology ($80.4\%$ F1). Comprehensive ablation studies evaluating key architectural components are provided in Appendix~\ref{sec:ablation}.

Unlike existing baselines that assume fixed node bounds, \ours{} dynamically predicts graph cardinality. As detailed in Table~\ref{tab:our:size_error}, the resulting node-count errors remain minimal across all datasets (e.g., $0.00$ on QM9 and $0.23$ on AIDS). To strictly penalize dimension mismatches, predicted or target adjacency matrices with incorrect sizes are zero-padded prior to computing the F1 score.

\begin{table}[t!]
\centering
\begin{tabular}{l|r@{}c@{}l|r@{}c@{}l|r@{}c@{}l|r@{}c@{}l}
Dataset$\backslash$method
    & \multicolumn{3}{c|}{PIGVAE [\%]}
    & \multicolumn{3}{c|}{ReGAE [\%]}
    & \multicolumn{3}{c|}{GRALE [\%]}
    & \multicolumn{3}{c}{\ours{} [\%]} \\
\hline
MUTAG      & \res{43.5}{14.9} & \res{60.1}{3.5}       & \res{6.9}{3.8}       & \best{61.3}{5.4} \\ 
AIDS       & \res{60.2}{4.9}    & \res{83.6}{2.6}         & \res{1.2}{0.6}   & \best{84.1}{1.7} \\
IMDB-BIN   & \res{87.3}{2.0}    & \res{90.0}{2.6}         & \res{54.8}{7.7}  & \best{94.0}{1.0} \\
QM9        & \res{0.0}{0.0}     & \best{99.9}{0.0}       & \res{98.6}{0.5}  & \res{99.4}{0.2}  \\
SYNTHETIC NEW  & \res{0.4}{0.6}     & \res{16.1}{3.0}         & \res{7.3}{0.7}   & \best{50.1}{0.4} \\
COLLAB     & \res{52.2}{5.0}    & \res{78.0}{2.4\,\dag}   & \oom               & \best{90.9}{0.7} \\
REDDIT-BIN & \oom               & \res{53.0}{2.0\,\ddag}  & \oom               & \best{56.5}{0.5} \\
\end{tabular}
\caption{Comparison of methods: F1 score on the test set. After $\pm$ we put the standard deviation. \dag  Score from 3 seeds, other 2 got NaN loss. \ddag  Score taken as reported in the original paper, due to training~instability.}
\label{tab:test_f1}
\vspace{1em}

\centering
\!\!\!\begin{tabular}{l|l|r@{}c@{}l|r@{}c@{}l|r@{}c@{}l|r@{}c@{}l|r@{}c@{}l}
Method & Dataset
    & \multicolumn{3}{c|}{\!\!Topo F1 [\%]\!\!}
    & \multicolumn{3}{c|}{\!\!Node F1 [\%]\!\!}
    & \multicolumn{3}{c|}{\!\!Node RMSE\!\!}
    & \multicolumn{3}{c|}{\!\!Edge F1 [\%]\!\!}
    & \multicolumn{3}{c}{\!\!Edge RMSE\!\!} \\
\hline
\multirow{3}{*}{PIGVAE}
    & MUTAG\! & \res{18.0}{9.7}         & \res{66.3}{9.5}      & \na                    & \best{71.2}{22.8}     & \na     \\
    & AIDS  & \res{41.9}{2.9}         & \res{56.7}{14.6}     & \res{1.5}{0.3}         & \res{77.7}{6.3}      & \na     \\
    & QM9   & \res{25.6}{15.9}       & \na                  & \best{0.5}{0.0}        & \na                  & \best{0.1}{0.0} \\
\hline
\multirow{3}{*}{GRALE}
    & MUTAG\! & \res{20.1}{18.3}      & \best{76.6}{25.1}   & \na                    & \res{56.3}{15.7}   & \na     \\
    & AIDS  & \res{0.1}{0.1}        & \res{8.3}{1.1}         & \res{4.3}{0.3}           & \res{35.5}{1.1}         & \na     \\
    & QM9   & \best{94.1}{0.7\,\dag} & \na                  & \res{5.4}{0.0\,\dag} & \na                  & \res{6.8}{0.3\,\dag} \\
\hline
\multirow{3}{*}{\ours{}\!}
    & MUTAG\! & \best{62.3}{3.0}       & \best{75.5}{5.9}    & \na                    & \res{58.7}{2.8}    & \na     \\
    & AIDS  & \best{80.4}{2.6}       & \best{62.2}{5.3}    & \best{0.8}{0.1}       & \best{85.0}{1.7}    & \na     \\
    & QM9   & \res{91.3}{1.2}            & \na                  & \best{0.5}{0.0}           & \na                  & \res{0.3}{0.1} \\
\end{tabular}

\caption{Comparison of methods: Topological, Node Features, and Edge Features reconstruction quality on the test set. F1 (higher is better) scores the categorical (one-hot) part of a feature block; RMSE (lower is better) scores its continuous part. N/A indicates that a metric is not applicable due to the nature of the features in a given dataset: F1 is N/A for purely continuous features (e.g., QM9 nodes and edges), while RMSE is N/A for purely categorical features (e.g., MUTAG node features, and edge features for MUTAG and AIDS). After $\pm$ we put the standard deviation. $\dagger$ denotes score obtained from 4 seeds, 1 seed got NaN loss.} \label{tab:test_f1_features}
\end{table}

\begin{table}  
\centering
\begin{tabular}{l|r@{${}\pm{}$}l|l|r@{${}\pm{}$}l}
Dataset & \multicolumn{2}{c|}{Mean size error} & Dataset & \multicolumn{2}{c}{Mean size error} \\
\hline 
MUTAG           & $0.12$  & $0.08$ &
AIDS            & $0.23$  & $0.09$ \\
IMDB-BIN        & $0.27$  & $0.05$ &
QM9             & $0.01$  & $0.00$ \\
SYNTHETIC NEW       & $0.00$  & $0.00$ &
COLLAB          & $1.45$  & $0.34$ \\
REDDIT-BIN      & $16.25$ & $1.50$ &
\end{tabular} 
\caption{GeoGAE: Mean size error --- the average absolute size difference between the target and predicted graphs over the average target graph size. Calculated for experiments for pure graph topology deconstruction, without features encoding and decoding enabled.
} 
\label{tab:our:size_error}
\vspace{-1em}
\end{table}

\section{Conclusions} 
\label{sec:conclusions} 

In this paper, we introduced \ours{}, a scalable graph-level autoencoder grounded in a novel hyperball cloud representation. By mapping discrete topologies into continuous geometric space with deterministic node ordering, \ours{} resolves the longstanding node-matching bottleneck and size limitations of existing models. Integrated with a Transformer architecture, our framework successfully compresses variable-sized graphs into fixed-dimensional vectors $z \in \mathbb{R}^m$ while enabling high-fidelity topological reconstruction. Across diverse benchmarks, \ours{} matches or exceeds state-of-the-art performance on most datasets, demonstrating unprecedented scalability on large graphs where prior methods fail. This establishes a robust foundation for latent-space graph generation, transformation, and optimization.


\paragraph{Limitations.}

Mapping a discrete graph to a hyperball cloud requires solving a continuous optimization problem  per graph, which adds a pre-processing overhead. This, however is done only once per dataset, and can be considered a new node's feature.
While our canonical ordering effectively mitigates the permutation ambiguity for most topologies, highly symmetrical graphs can remain sensitive to minor spatial perturbations during SVD normalization and greedy ordering.
Scaling to large graphs introduces challenges in precise graph size prediction and topology preservation under node bundling. Addressing end-to-end differentiable hyperball initialization and symmetry-aware ordering remains a promising direction for future work.

\subsubsection*{Reproducibility Statement}

We have made all efforts to ensure the full reproducibility of both our theoretical claims and empirical results.

\textbf{Theoretical Reproducibility:} 
The mathematical guarantee that any graph can be losslessly represented as a set of intersecting hyperballs is formally stated in Proposition \ref{proposition1}, with a constructive proof provided in Appendix \ref{proof}. The deterministic node ordering mechanism is explicitly detailed in Algorithm \ref{alg:order} (Appendix \ref{node_ordering}).

\textbf{Empirical Reproducibility:}
\begin{itemize}
    \item \textbf{Hyperparameters \& Training:} All hyperparameters for GeoGAE and the baseline methods (PIGVAE, ReGAE, GRALE) are comprehensively documented in Tables \ref{tab:our:hyperparameters}, \ref{tab:pigvae:hyperparameters}, \ref{tab:regae:hyperparameters}, and \ref{tab:grale:hyperparameters}, respectively. We explicitly report batch sizes, embedding dimensions, network depths, learning rates, and weight decays.
    \item \textbf{Compute \& Infrastructure:} The exact hardware specifications (including NVIDIA A100 GPUs and AMD EPYC processors) and software environments utilized across four computational clusters are detailed in Appendix \ref{app:hardware} (Table \ref{tab:hardware}). To aid in reproducing our training pipelines, we report the estimated end-to-end execution times per dataset on a single A100 GPU in Table \ref{tab:hardware_feature_decoding} (e.g., $\sim$38 hours for QM9 with feature decoding).
    \item \textbf{Codebase:} The complete source code, environment specifications, and training scripts with execution instructions are provided in the supplementary materials.
\end{itemize}

\subsubsection*{LLM Usage Statement}
\label{sec:llm_usage}

In accordance with the ICLR 2027 Policy on AI Assistance, we disclose the use of Large Language Models (LLMs) during the preparation of this work:
\begin{itemize}
\item \textbf{Writing and Editing:} LLMs were utilized as writing assistants to refine grammar, improve sentence flow, format LaTeX tables, and ensure stylistic consistency across the manuscript.

\item \textbf{Literature Discovery:} LLMs were employed to assist in discovery and literature search for relevant baseline methodologies and related work in geometric deep learning. All retrieved citations and statements were manually verified for accuracy by the authors.

\item \textbf{Code Execution \& Implementation:} LLMs assisted in the implementation phase by suggesting code refactoring for PyTorch modules, generating boilerplate code for data loading, and aiding in the debugging of minor implementation details.
\end{itemize}

\noindent\textbf{Author Responsibility:} We emphasize that no LLMs were used to generate the core scientific contributions, mathematical proofs (including Proposition 1), or conceptual designs. All baseline comparisons, experimental evaluations, and final interpretations were exclusively conceived, executed, and verified by the authors. The authors take full responsibility for the original content, code correctness, and scientific validity of this article.





\subsubsection*{Acknowledgments}
We gratefully acknowledge the Polish high-performance computing infrastructure PLGrid (HPC Center: ACK Cyfronet AGH) for providing computational resources and support under grant no.~PLG/2025/018560 and PLG/2026/019802.

\ifdefined\arxivversion

\else    \bibliographystyle{iclr2027_conference}
    \bibliography{references,references2,references3}
\fi

\appendix

\section{Proof of Proposition 1}
\label{proof}

The proof is by construction. The radius of each ball is $0.5\sqrt{2n-1}$. The center of each ball is a~vector of $0$s and exactly $n$ $1$s. If two nodes are adjacent, they share $1$ at a~single coordinate; if they are not, they do not share any~$1$s; only two balls may share $1$ at the same coordinate. The center of the first ball is $[1,\dots,1,0,\dots,0]$. The center of $i$-th ball is constructed as follows. For adjacent balls $1,\dots,i-1$, we copy their $1$s at lowest coordinates at which they do not share~$1$ with other balls. The remaining $1$s are set at coordinates starting from $(i\cdot n+1)$-th. 

For this coordination we need at most $n^2$ coordinates. 

If the nodes are adjacent, the distance between their corresponding ball centers is $\sqrt{2n-2}$, thus they intersect. If they are not adjacent, the distance is $\sqrt{2n}$, thus they do not intersect. \qed{}

For more results on graph sphericity, see \citep{1983fishburn,1986maehara}. 

\section{Details of the \ours{} architecture} 

\subsection{Ordering nodes}
\label{node_ordering}

Algorithm~\ref{alg:order} presents details of the node ordering. Note that due to normalization of embeddings,~$\bar x=\mathbf{0}$. 

\begin{algorithm}[h]
\caption{Ordering nodes}
\label{alg:order} 
\begin{algorithmic}[1]
\STATE {\bfseries Input:} Geometric embeddings $(x^g_1, \dots, x^g_n)$ where $x^c_j$ and $r_j$ denote the ball center and radius of the $j$-th node.
\STATE $\bar x \leftarrow \frac{1}{n} \sum_{k=1}^n x^c_k$
\STATE $\pi(1) \leftarrow \arg\min_{j\in\{1,\dots,n\}} \|x^c_j - \bar x\|$
\STATE $y^g_1 = x^g_{\pi(1)}$
\STATE $\setA \leftarrow \{1,\dots,n\} \setminus \{\pi(1)\}$
\FOR{$i=2,\dots,n$} 
    \STATE $\pi(i) \leftarrow \arg\min_{j\in \setA} \left( \frac{\|x^c_j - x^c_{\pi(i-1)}\|}{0.5(f(r_j) + f(r_{\pi(i-1)}))} + \frac{\log_2 \|x^c_j - \bar x\|}{10} \right)$
    \STATE $y^g_i = x^g_{\pi(i)}$
    \STATE $\setA \leftarrow \setA \setminus \{\pi(i)\}$
\ENDFOR
\STATE {\bf return} $(y^g_1, \dots, y^g_n)$ 
\end{algorithmic}
\end{algorithm}

\subsection{Formal Mechanics of Node Bundling}
\label{app:bundles}
Given an ordered sequence of node embeddings $Y = (y_1^g, \dots, y_n^g) \in \mathbb{R}^{n \times d_{\text{geom}}}$, the bundling procedure operates as follows:
\begin{enumerate}
    \item \textbf{Padding:} If $n \pmod b \neq 0$, $n_{\text{pad}} = b - (n \pmod b)$ copies of the final node $y_n^g$ are appended to $Y$, yielding a padded sequence of length $n' = \lceil n/b \rceil \cdot b$.
    \item \textbf{Reshaping \& Projection:} The padded matrix is reshaped into $T = n'/b$ bundle blocks of shape $[T, b \cdot d]$ and mapped via an MLP, $\phi: \mathbb{R}^{b \cdot d} \to \mathbb{R}^{d_{\text{tok}}}$, to~form the encoder sequence tokens $[t_1, \dots, t_T]$.
    \item \textbf{Decoding:} In the decoder path, each generated token is mapped via an MLP, $\decod: \mathbb{R}^{d_{\text{tok}}} \to \mathbb{R}^{b \cdot d} \times \mathbb{R}^b$, into $b$ raw geometric embeddings alongside $b$ stopping logits used to accurately truncate any padded nodes in the final bundle.
\end{enumerate}

\subsection{Masking encoder tokens}
\label{app:encoder_masking}
\begin{figure}
    \centering
    \resizebox{\columnwidth}{!}{
        \definecolor{cInput}{HTML}{4C72B0}
\definecolor{cBundle}{HTML}{55A868}
\definecolor{cCls}{HTML}{DD8452}
\definecolor{cTf}{HTML}{8172B3}
\definecolor{cLatent}{HTML}{C44E52}
\definecolor{cAttn}{HTML}{4E79A7}
\definecolor{cFfn}{HTML}{59A14F}
    
    \begin{tikzpicture}[
        font=\large,
        >={Stealth[round]},
        vtx/.style   ={circle, draw=cInput!85!black, fill=cInput!22, semithick,
            minimum size=11mm, inner sep=0pt, font=\large},
        clsb/.style  ={rounded corners=2.5pt, draw=cCls!85!black, fill=cCls!28,
            semithick, minimum width=14mm, minimum height=10mm, inner sep=2pt},
        nn/.style    ={draw=black!38, semithick},                
        cn/.style    ={draw=cCls!65, semithick, opacity=0.75},   
        block/.style ={draw=black!45, semithick},
        cap/.style   ={font=\Large, black!55},
        tag/.style   ={font=\Large, black!60},
        head/.style  ={font=\Large\bfseries, black!70},
        ]
        
        \node[clsb] (c1) at (0.4, 1.2) {\large[CLS]$_1$};
        \node[clsb] (c2) at (0.4,-1.2) {\large[CLS]$_2$};
        
        \node[vtx] (v1) at (4.3, 1.75) {$v_1$};
        \node[vtx] (v2) at (6.3, 1.75) {$v_2$};
        \node[vtx] (v3) at (6.3,-1.75) {$v_3$};
        \node[vtx] (v4) at (4.3,-1.75) {$v_4$};
        
        \begin{scope}[on background layer]
            \draw[nn] (v1)--(v2); \draw[nn] (v2)--(v3); \draw[nn] (v3)--(v4);
            \draw[nn] (v4)--(v1); \draw[nn] (v1)--(v3); \draw[nn] (v2)--(v4);
            \foreach \v in {v1,v2,v3,v4}{
                \draw[cn] (c1)--(\v);
                \draw[cn] (c2)--(\v);
            }
        \end{scope}
        
        \draw[cLatent!75!black, dashed, semithick] (c1)--(c2);
        \node[circle, draw=cLatent!80!black, fill=white, thick, inner sep=1.2pt,
        font=\small, text=cLatent!70!black] (xm) at (0.4,0) {$\times$};
        \node[tag, text=cLatent!70!black, right=2pt of xm] {$-\infty$};
        
        \begin{scope}[on background layer]
            \node[draw=cInput!45, dashed, rounded corners=10pt, inner sep=9pt,
            fit=(v1)(v2)(v3)(v4)] (vhalo) {};
        \end{scope}
        \node[tag, text=cInput!45!black] at (5.3,3.05) {graph vertices ($v_1,\dots,v_T$)};
        \node[tag, text=cCls!45!black, align=center] at (0.4,2.95) {learned\\ {[}CLS{]} tokens};
        \node[head] at (3.3,3.95) {allowed attention (edges)};
        
        \node[tag, text=black!45] at (5.3,0) {full};
        
        \begin{scope}[shift={(-0.8,-3.5)}]
            \draw[cn]  (0,0.0)  -- ++(0.9,0);
            \node[tag, right=3pt] at (0.9,0.0) {[CLS] $\leftrightarrow$ every vertex};
            \draw[nn]  (0,-0.7) -- ++(0.9,0);
            \node[tag, right=3pt] at (0.9,-0.7) {vertex $\leftrightarrow$ vertex (all pairs)};
            \draw[cLatent!75!black, dashed, semithick] (0,-1.4) -- ++(0.9,0);
            \node[circle, draw=cLatent!80!black, fill=white, thick, inner sep=0.8pt,
            font=\small, text=cLatent!70!black] at (0.45,-1.4) {$\times$};
            \node[tag, right=3pt] at (0.9,-1.4) {[CLS] $\leftrightarrow$ [CLS]: masked ($-\infty$)};
            \node[tag, black!45, right=3pt] at (-0.25,-2.1)
            {edges are bidirectional; self-attention allowed};
        \end{scope}
        
        \begin{scope}[shift={(12.5,2.3)}]      
            \def\C{1.2}                          
            \def\N{3.4}                          
            \pgfmathsetmacro{\W}{\C+\N}          
            \def\h{0.6}                          
            
            \fill[cFfn!16] (\C,0) rectangle (\W,-\C);            
            \fill[cFfn!16] (0,-\C) rectangle (\C,-\W);           
            \fill[cFfn!20] (\C,-\C) rectangle (\W,-\W);          
            
            \fill[cFfn!16]  (0,0)      rectangle (\h,-\h);        
            \fill[cFfn!16]  (\h,-\h)   rectangle (\C,-\C);        
            \fill[cLatent!35] (\h,0)    rectangle (\C,-\h);       
            \fill[cLatent!35] (0,-\h)   rectangle (\h,-\C);       
            \node[font=\small,text=cLatent!45!black] at (\h+0.3,-0.3) {$-\infty$};
            \node[font=\small,text=cLatent!45!black] at (0.3,-\h-0.3) {$-\infty$};
            
            \node[font=\large] at (\C+\N/2,-\C-\N/2) {$\mathbf{0}$};
            \node[tag, text=black!45] at (\C+\N/2,-\C-\N/2-0.5) {full attention};
            \node[font=\large] at (\C+\N/2,-\C/2) {$\mathbf{0}$};
            \node[font=\large] at (\C/2,-\C-\N/2) {$\mathbf{0}$};
            
            \draw[block] (0,0) rectangle (\W,-\W);
            \draw[cCls!75!black, thick] (\C,0.0) -- (\C,-\W-0.0);
            \draw[cCls!75!black, thick] (-0.0,-\C) -- (\W+0.0,-\C);
            \draw[black!25,thin] (\h,0) -- (\h,-\C);
            \draw[black!25,thin] (0,-\h) -- (\C,-\h);
            
            \draw[decorate,decoration={brace,amplitude=5pt},black!55]
            (0,0.18) -- (\C,0.18) node[midway,above=5pt,tag, font=\large]{CLS ($C$)};
            \draw[decorate,decoration={brace,amplitude=5pt},black!55]
            (\C,0.18) -- (\W,0.18) node[midway,above=5pt,tag, font=\Large]{vertices ($T$)};
            \draw[decorate,decoration={brace,amplitude=5pt,mirror},black!55]
            (-0.18,0) -- (-0.18,-\C);
            \draw[decorate,decoration={brace,amplitude=5pt,mirror},black!55]
            (-0.18,-\C) -- (-0.18,-\W);
            \node[tag,rotate=90, font=\large] at (-0.8,-\C/2)    {CLS};
            \node[tag,rotate=90] at (-0.8,-\C-\N/2) {vertices};
            
            \node[head] at (\W/2,1.7) {additive mask};
            \node[tag, text=black!50] at (\W/2+0.55,-\W-1.2)
            {keys $\rightarrow$ (columns),\; queries $\downarrow$ (rows)};
        \end{scope}
        
        \begin{scope}[shift={(11.7,-4.5)}]
            \fill[cFfn!20,draw=black!35] (0,0) rectangle (0.45,0.45);
            \node[tag,right=3pt] at (0.45,0.225) {$0$ (attend)};
            \fill[cLatent!35,draw=black!35] (0.0,-0.7) rectangle (0.45,-0.25);
            \node[tag,right=3pt] at (0.45,-0.475) {$-\infty$ (masked)};
        \end{scope}
        
        \node[font=\Huge, black!45] at (9.4,0.2) {$\equiv$};
        
    \end{tikzpicture}
    }
    \caption{Encoder self-attention mask. The token sequence is
        $[\,\texttt{[CLS]}_1,\dots,\texttt{[CLS]}_C,\,v_1,\dots,v_T\,]$. Each
        \texttt{[CLS]} token attends to every graph vertex and every vertex attends
        back (global read/write), and vertices attend to one another over the
        complete set of pairs, so node--node attention is fully connected and
        permutation-invariant. The only forbidden interactions
        are between distinct \texttt{[CLS]} tokens, which are removed with an additive
        $-\infty$ entry; self-attention on the diagonal is kept. Left: the allowed
        attention edges. Right: the equivalent additive mask, whose
        $\texttt{[CLS]}\times\texttt{[CLS]}$ off-diagonal block is masked while all
        \texttt{[CLS]}--vertex and vertex--vertex blocks are zero.}
    \label{fig:attention-mask}
\end{figure}
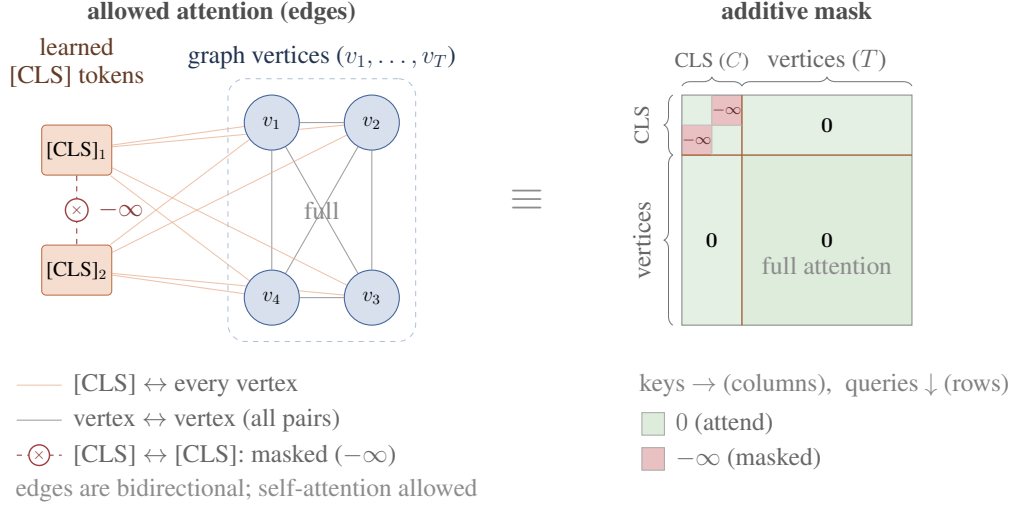

Details of masking in the encoder attention are presented in Figure~\ref{fig:attention-mask}. We make sure class tokens do not attend to each other -- this results in each of them is forced to learn a meaningful information about the rest of the tokens in a different way. Intuitively, the encoder may learn to transform different class tokens into orthogonal characteristics of the graph. 

\subsection{Optimization objective}

The total objective function optimized during training is a combination of: a geometric reconstruction loss, hyperball reconstruction loss, an end of output sequence (EOS) loss, an optional Kullback-Leibler (KL) divergence regularization, and optional node and edge features' reconstruction losses. Crucially, to mitigate exposure bias (as described in Section~\ref{sec:ML-TF}), the losses are accumulated over all $K$ iterative layers of the decoder.

Let $\hat{x}^{(k)}_i=[\hat x_i^{c,(k)},\hat r_i^{(k)}]$ denote the geometric node embedding predicted at the $k$-th decoder layer, split into its center and raw radius as in Section~\ref{sec:ball-cloud}, and let
\begin{equation} \label{def:m^(k)}
m_{i,j}^{(k)} = 0.75\big(f(\hat r_i^{(k)})+f(\hat r_j^{(k)})\big) - \|\hat x_i^{c,(k)}-\hat x_j^{c,(k)}\|
\end{equation}
be the predicted margin between nodes $i$ and $j$. We treat edge existence as a binary classification problem on $p_{i,j}^{(k)}=\sigma(m_{i,j}^{(k)}/T)$ against $A_{i,j}$, $T>0$ is constant, using the same focal-loss construction as in Eq.~\eqref{opt:x} ($\widehat p_{ij}^{(k)},\alpha_t$ defined analogously, substituting $A_{i,j}$ for $D_{i,j}=1$):
\begin{equation}
    \mathcal{L}_{\text{geo}}^{(k)} = \frac{1}{\binom{n}{2}}\sum_{1\leq i<j\leq n} -\alpha_t\,(1-\widehat p_{ij}^{(k)})^\gamma\log(\widehat p_{ij}^{(k)}).
\end{equation}

To improve the reconstruction of the original embeddings to preserve the structure of the generated embeddings, we also concurrently use the Huber loss:
\begin{equation} \label{Huber:loss}
    \mathcal{L}_{\text{emb}}^{(k)} = \frac{1}{n}\sum_{i=1}^n H_\delta\big(\hat{x}_i^{(k)} - x_i\big), \qquad
    H_\delta(z) = \begin{cases}
                      \frac{1}{2}z^2 & |z| \le \delta \\
                      \delta\left(|z| - \frac{1}{2}\delta\right) & \text{otherwise}
    \end{cases}
\end{equation}  

Concurrently, the model must accurately predict the graph size. The decoder predicts stop-token logits for each generated node slot. We treat this as a binary classification problem where the ground-truth target $y_i \in \{0, 1\}$ is $1$ if the slot index meets or exceeds the actual graph size, and $0$ otherwise. Given the significant class imbalance (mostly non-stop tokens), we evaluate the predictions using Focal Loss. Following the standard notation, let $q_i \in [0,1]$ be the model's estimated probability for the $i$-th token being a stop token. We formally define $\widehat q_{i}$ and $\eta_i$ as:
\begin{equation}
    \widehat q_{i} = \begin{cases}
                         q_i & \text{if } y_i = 1 \\
                         1 - q_i & \text{otherwise}
    \end{cases}, \quad
    \eta_i = \begin{cases}
                   \eta & \text{if } y_i = 1 \\
                   1 - \eta & \text{otherwise}
    \end{cases},
\end{equation}
where $\eta \in [0, 1]$ is a weighting factor for the positive class. The stopping loss is then concisely formulated as:
\begin{equation}
    \mathcal{L}_{\text{stop}}^{(k)} = \frac{1}{n} \sum_{i=1}^n -\eta_i (1 - \widehat q_{i})^\gamma \log(\widehat q_{i}),
\end{equation}
with $\gamma \geq 0$ serving as a focusing parameter that smoothly down-weights the loss for easily classified examples.

Optionally, when the model is configured to decode node, and edge features, we use the following feature reconstruction losses:
\begin{equation}
    \mathcal{L}_{\text{nfeat}}^{(k)} = \frac{1}{n}\sum_{i=1}^n H_1(\hat c_i^{(k)} - c_i) + \mathrm{CE}(\hat{\bar c}_i^{(k)},\ \bar c_i)
\end{equation}
\begin{equation}
    \mathcal{L}_{\text{efeat}} = \frac{1}{|E|}\sum_{(i,j)\in E} H_1(\hat e_{ij}^{(k)} - e_{ij}) + \mathrm{CE}(\hat {\bar e}_{ij}^{(k)},\ \bar e_{ij})
\end{equation}
where $c_i$ $(e_{ij})$ and $\hat c_i^{(k)}$ $(\hat e^{(k)}_{ij})$ are continuous node (edge) features and their reconstructions, $\bar c_i$ $(\bar e_{ij})$ and $\hat {\bar c}^{(k)}_i$ $(\hat {\bar e}^{(k)}_{ij})$ are 1-hot encodings of discrete node (edge) features and their reconstructions, $H_1$ is the Huber loss \eqref{Huber:loss}, and $\mathrm{CE}$ is the standard Cross-Entropy loss. 

The total loss for a single graph is given by an averaging the geometric embedding reconstruction loss $\mathcal{L}_{\text{emb}}^{(k)}$ altogether with the geometric and stopping losses across all $K$ instances of the decoder, and adding the KL divergence $\mathcal{L}_{\text{KL}}$ between the approximate posterior and the standard Gaussian prior, weighted by a linear warmup $\beta(t)$ from $0$ to a target scale over the first training epochs:
\begin{equation}
    \mathcal{L}_{\text{total}} = \beta(t)\,\mathcal{L}_{\text{KL}} + \frac{1}{K}\sum_{k=1}^K \left( \mathcal{L}_{\text{emb}}^{(k)} + \rho \mathcal{L}_{\text{geo}}^{(k)} + \mathcal{L}_{\text{stop}}^{(k)} \right) + \lambda_{\text{feat}}\left( \sum_{k=1}^K \mathcal{L}_{\text{nfeat}}^{(k)} + \mathcal{L}_{\text{efeat}} \right).
\end{equation}
We divide the presented losses into: KL-div loss: $\mathcal{L}_{\text{KL}}$, topological losses: $\mathcal{L}_{\text{emb}}^{(k)}, \mathcal{L}_{\text{geo}}^{(k)},\mathcal{L}_{\text{stop}}^{(k)}$, and feature reconstruction losses: $\mathcal{L}_{\text{nfeat}}, \mathcal{L}_{\text{efeat}}$. When VAE is disabled, the loss: $\mathcal{L}_{\text{KL}}$ is not used; When features decoding is disabled, feature losses: $\mathcal{L}_{\text{nfeat}}, \mathcal{L}_{\text{efeat}}$ are also not present in the $\mathcal{L}_{\text{total}}$. $\rho$ is a~hyperparameter. 

In practice, each component of the topological loss is additionally rescaled every batch to match the
mean loss magnitude across components (capped) before the static weights above
are applied, so that no single term dominates the gradient purely by raw
scale. During batch
processing, $\mathcal{L}_{\text{total}}$ is averaged across all graphs in the
batch.

\section{Benchmark datasets}

The datasets used in the experimental study are MUTAG \citep{debnath1991structure}, AIDS \citep{riesen2008iam}, IMDB-BINARY \citep{kriege2020survey}, 
QM9 \citep{Ramakrishnan2014QM9}, and SYNTHETIC NEW \citep{Morris2020TUDataset}, COLLAB, REDDIT-BINARY  \citep{yanardag2015deep}. They are characterized in Table~\ref{tab:datasets:app}. 

\begin{table*}
\begin{tabular}{p{1.8cm}|p{5.3cm}|p{1.1cm}|p{1.2cm}|p{2.4cm}}
Dataset & Domain and graph construction &	Task & Classes\,/ targets &	Provided node/edge information \\ 
\hline
MUTAG & 
Molecular compounds; mutagenicity classification & 
Graph classification & 
2 classes & 
Nodes: 7 discrete atom types; edges: 4 bond types \\ 
AIDS & 
Molecular compounds; anti-HIV activity classification & 
Graph classification & 
2 classes & 
Nodes: discrete labels + 4 continuous attributes; edges: 4 bond types \\ 
IMDB-BINARY & 
Movie collaboration ego-networks. Nodes are actors; an edge connects actors appearing in the same movie. Graphs come from Action and Romance movies. &  
Graph classification &
2 classes &
No native node labels, node attributes, edge labels, or edge attributes \\ 
QM9 & 
Small organic molecules. Nodes are atoms and edges are chemical bonds; optimized 3-D coordinates and quantum-chemical properties are supplied. Molecules contain H, C, N, O and F, with at most nine non-hydrogen atoms. & 
Graph-level regression & 
Multiple continuous targets & 
Atom types/features, bond types and 3-D coordinates \\ 
SYNTHE\-TIC NEW & 
Artificial graphs designed for controlled graph-classification experiments. Each graph has exactly 100 vertices and approximately 196 edges. & 
Graph classification & 
2 classes & 
Discrete node labels and one-dimensional continuous node attributes; no edge labels \\
COLLAB  & 
Scientific collaboration ego-networks. Nodes are researchers and edges represent co-authorship. Graph labels correspond to three physics research fields. & 
Graph classification & 
3 classes & 
No native node or edge features/labels \\ 
REDDIT-BINARY &
Reddit discussion graphs. Nodes represent users; edges represent interactions/replies in a discussion thread. The two classes group discussions from different types of subreddits.	& 
Graph classification  &
2 classes & 
No native node or edge features/labels 
\end{tabular}
\caption{Datasets}
\label{tab:datasets:app}
\end{table*}


\section{Hardware and Computational Runtime}
\label{app:hardware}

All experiments and hyperparameter optimizations were conducted across high-performance computing infrastructure, primarily leveraging two computational allocation grants on the supercomputer equipped with NVIDIA A100 GPUs (80GB). Due to job scheduling limits and queue constraints on the primary cluster, secondary computational environments (detailed in Table~\ref{tab:hardware}) were utilized in parallel to distribute the workload.

Across the entire research lifecycle—encompassing preliminary model explorations, hyperparameter sweeps, baseline evaluations, and ablation studies—the total compute expenditure accumulated to approximately 32,000 GPU-hours (20,000 and 12,000 GPU-hours across the two respective allocation grants).
To ensure full transparency and reproducibility, the estimated end-to-end execution times required to complete a full benchmark evaluation (averaged over 5 independent random seeds per dataset) on a single NVIDIA A100 GPU are summarized in Table \ref{tab:hardware_feature_decoding}.

\begin{table*}
    \centering
    \begin{tabular}{|p{1.5cm}|p{1.5cm}|p{1.5cm}|p{2cm}|p{3cm}|p{1cm}|}
        \hline
        \textbf{Cluster name} & \textbf{GPU device} & \textbf{GPU driver version} & \textbf{CPU device} & \textbf{Operating system} & \textbf{Total RAM}  \\
        \hline
        \textbf{Cluster 1} &  NVIDIA A100 PCIe 80GB & 565.57.01 & AMD EPYC 7713 64-Core Processor & Linux-6.8.0-64-generic-x86\_64-with-glibc2.39 & 2048 GB \\
        \hline
        \textbf{Cluster 2} & DGXA100 920-23687-2531-001 & 580.173.02 & AMD EPYC 7742 64-Core Processor & Linux-5.15.0-1107-nvidia-with-glibc2.34 & 1024 GB\\
        \hline
        \textbf{Cluster 3} & AMD Radeon RX 9070 XT (Navi 48) & Mesa 26.2.2 / ROCm 7.8.0 & AMD Ryzen 9 9950X3D 16-Core Processor & Linux-6.18.49-1-MANJARO-x86\_64-with-glibc2.44 & 64 GB \\
        \hline
        \textbf{Cluster 4} &  NVIDIA RTX PRO 500 Blackwell Generation & 580.173.02 & Intel(R) Core(TM) Ultra 5 235H 14-Core Processor & Linux-7.0.0-28-generic-x86\_64-with-glibc2.39 & 30 GB \\
        \hline
    \end{tabular}
    \caption{Specification of the computational clusters used in the experiments.}
    \label{tab:hardware}
\end{table*}

\begin{table}[htbp]
\centering
\begin{tabular}{l|c|c}
\toprule
\textbf{Dataset} & \textbf{No features} & \textbf{With feature decoding} \\
\midrule
MUTAG          & $\approx 24\text{ min}$ & $\approx 25\text{ min}$ \\
AIDS           & $\approx 2\text{h } 55\text{ min}$ & $\approx 18\text{h } 4\text{ min}$ \\
IMDB-BIN       & $\approx 1\text{h } 50\text{ min}$ & N/A \\
SYNTHETIC NEW  & $\approx 44\text{ min}$ & N/A \\
COLLAB         & $\approx 23\text{h } 37\text{ min}$ & N/A \\
QM9            & $\approx 29\text{h } 16\text{ min}$ & $\approx 38\text{h } 16\text{ min}$ \\
REDDIT-BIN     & $\approx 21\text{h } 39\text{ min}$ & N/A \\
\bottomrule
\end{tabular}
\caption{Runtime comparison for \our{} across datasets with and without feature decoding, calculated on \textbf{Cluster 1}. We report results averaged over 5 seeds. N/A denotes where computations were not applicable (datasets with no node or edge features).}
\label{tab:hardware_feature_decoding}
\end{table}

\section{Setup of GRALE, PIGVAE, ReGAE}

Authors of all three of the methods presented as baseline for \ours{}, have shared reference implementation of their algorithms, and these implementations were utilized in calculating their scores. For all four of the evaluated algorithms, the dataset reading, splitting, training loop, metrics calculation and reporting code was common to ensure fair reporting. The hyperparameters reported in Tables~\ref{tab:pigvae:hyperparameters}, \ref{tab:regae:hyperparameters} and \ref{tab:grale:hyperparameters} were determined using automatic parameter hypertuning algorithm, separately for each algorithm and dataset, only exceptions being REDDIT-BIN and COLLAB for which, due to limited computational capacity, parameters were taken from the original papers or tuned manually. The best results for PIGVAE and GRALE are obtained for relatively small graph embedding sizes. Surprisingly, for larger embedding sizes, the scores for these architectures are worse. 

\begin{table*} 
\centering
\begin{tabular}{l|r|r|r|r|r|r|r|r|r}
Dataset & emb & hid & ppf-h & heads & layers & k-ls & p-ls & lr   & w-decay \\
\hline 
MUTAG          & 64       & 128         & 512              & 4          & 4           & 1e-3             & 1e-1              & 1e-4 & 0.0     \\ 
AIDS          & 64       & 128         & 512              & 4          & 4           & 1e-3             & 1e-1              & 1e-4 & 0.0     \\
IMBD-BIN      & 16       & 256         & 512              & 4          & 4           & 1e-2             & 1e-1              & 1e-4 & 1e-4     \\
QM9           & 32       & 128         & 256              & 2          & 1           & 1e-3             & 5e-1              & 1e-4 & 0.0      \\
SYNTHETIC & 16       & 256         & 256              & 8          & 4           & 1e-2             & 1.0               & 1e-4 & 0.0      \\
COLLAB    & 32       & 128         & 256              & 2          & 1           & 1e-3             & 5e-1              & 1e-4 & 0.0      \\
REDDIT-BIN    & 32       & 128         & 256              & 2          & 1           & 1e-3             & 5e-1              & 1e-4 & 0.0    
\end{tabular} 
\caption{Hyperparameters of PIGVAE: emb -- graph embedding size, hid -- hidden layer size,  ppf-h -- transformer layer size, heads -- number of transformer heads, layers -- number of transformer encoder/decoder layers, k-ls -- kld loss scale, p-ls -- permutation loss scale, lr -- learning rate, w-decay -- weight decay.}
\label{tab:pigvae:hyperparameters} 
\end{table*} 

\begin{table*} 
\centering 
\begin{tabular}{l|r|r|r|r|r|r}
Dataset       & emb & encod & decod & block & l-r   & w-decay \\
\hline
MUTAG          & 160       & 2048                   & 2048                   & 4           & 3e-4 & 1e-3   \\ 
AIDS          & 160       & 2048                   & 2048                   & 4           & 3e-4 & 1e-3   \\ 
IMBD-BIN      & 160       & 2048                   & 4096                   & 4           & 5e-4 & 1e-4     \\
QM9           & 160       & 1024                   & 1024                   & 32          & 3e-4 & 1e-3     \\
SYNTHETIC & 160       & 2048                   & 4096                   & 8           & 1e-3 & 1e-3     \\
COLLAB & 604       & 2048, 1536                   & 4096                   & 16           & 3e-4 & 1e-3     \\
REDDIT-BIN    & 1720      & 4096                   & 6144                   & 64          & 1e-4 & 1e-4    
\end{tabular} 
\caption{Hyperparameters of ReGAE: emb -- embedding size, encod -- sizes of encoder hidden layers, decod -- sizes of decoder hidden layers, block -- block size, l-r -- learning rate, w-decay -- weight decay of optimizer. For all experiments we set 0.5 as the mask weight and 0.2 as the embedding norm weight for the loss calculation. We use ELU~\citep{2016djork} as the activation function.} 
\label{tab:regae:hyperparameters} 
\end{table*} 

\begin{table*}[t]
\centering
\setlength{\tabcolsep}{5pt}
\newcommand{\dims}[2]{#1 & $\times$ & #2}
\begin{tabular}{l|c|c|r@{}c@{}l|r@{}c@{}l|c|c|c|c|r}
Dataset & \!Layers\! & \!\!Heads\!\!
    & \multicolumn{3}{c|}{\!Node Dim\!}
    & \multicolumn{3}{c|}{Edge Dim}
    & \!\!Latent\!\! & \!\!Matcher\!\! & LR & \!\!Dropout\!\! & \!Batch\! \\
\hline
MUTAG        & 3    & 8   & \dims{64}{64}   & \dims{null}{64} & 256  & Soft  & 1e-4 & 0.0   & 64  \\
AIDS         & 7    & 8   & \dims{128}{128} & \dims{128}{128} & 128  & Soft  & 2e-5 & 0.1   & 8   \\
IMDB-BIN  & 3    & 4   & \dims{64}{64}   & \dims{32}{32}   & 64   & Sink  & 2e-5 & 0.0   & 4   \\
QM9          & 6    & 8   & \dims{64}{128}  & \dims{64}{64}   & 128  & Sink  & 1e-4 & 0.0   & 256 \\
SYNTH. NEW    & 5    & 8   & \dims{64}{64}   & \dims{32}{32}   & 64   & Sink  & 1e-4 & 0.0   & 8   \\
COLLAB       & 3    & 4   & \dims{64}{128}  & \dims{64}{64}   & 64   & Sink  & 1e-4 & 0.0   & 1   \\
REDDIT-BIN   & 2    & 2   & \dims{64}{128}  & \dims{64}{64}   & 128  & Sink  & 1e-4 & 0.0   & 1   \\
\end{tabular}
\caption{Hyperparameters of GRALE: Layers -- number of Evoformer layers, Heads -- attention heads, Node/Edge Dim -- hidden $\times$ model dimension, Latent -- graph embedding dimension ($d_g$), Matcher -- node matching operator (Sink: Sinkhorn, Soft: SoftSort), LR -- learning rate, Dropout -- attention/MLP dropout, Batch -- batch size.}
\label{tab:grale:hyperparameters}
\end{table*}

\section{Ablation study}
\label{sec:ablation}
In this section, we conduct a comprehensive ablation study to evaluate the impact of key architectural choices and training strategies on GeoGAE's ability to reconstruct graph topologies. We evaluate these variations on three datasets containing both node and edge features: MUTAG, AIDS, and QM9. The quantitative results of these experiments are summarized in Table \ref{tab:ablation_study} and visually supported by Figure \ref{fig:tf-ablation}, and Figure \ref{fig:bundle-size-ablation}.\paragraph{Multi-layer Teacher Forcing and Exposure Bias.}
Autoregressive Transformer decoders are notoriously susceptible to exposure bias during inference, as they must rely on their own potentially flawed predictions instead of ground-truth tokens. We hypothesized that our multi-layer teacher forcing strategy (stacking the decoder $K$ times) mitigates this issue. To isolate its effect, we evaluated the model using only a single decoder iteration ($K=1$) and compared it against our default setting ($K=3$). As shown in Table \ref{tab:ablation_study}, reducing the iterations to one causes a drastic collapse in performance. For instance, the topological F1 score on the AIDS dataset drops from $84.12\%$ to $72.59\%$. More importantly, the model completely loses its ability to correctly predict the target graph size, with the average size difference skyrocketing from $0.23$ to $4.67$. Figure \ref{fig:tf-ablation} further illustrates this phenomenon across different values of $K$ on the AIDS dataset. The reconstruction performance and size prediction stabilize strictly at $K=2$. This proves that a small number of iterative refinement steps is both crucial and sufficient to eliminate cascading errors during graph generation.\paragraph{Impact of Variational Regularization (VAE).}
Next, we investigated the effect of enabling the full Variational Autoencoder (VAE) formulation. Adding the Kullback-Leibler (KL) divergence loss forces the latent space to conform to a standard Gaussian prior, which is typically desired for generative sampling.However, our results indicate that this regularization slightly degrades the exact topology reconstruction capability. Across all three tested datasets, the purely deterministic model outperforms the VAE variant in reconstruction accuracy. For example, on the QM9 dataset, the topological F1 score decreases from $99.38\%$ to $95.88\%$ when VAE is enabled. This highlights a natural trade-off in graph autoencoders between achieving exact deterministic reconstruction (autoencoding) and maintaining a smooth, sampleable latent space (generation).\paragraph{Influence of Node and Edge Features on Topology.}
Also, we examined whether encoding node and edge attributes helps the model comprehend the underlying graph structure better, even when the decoder is not explicitly tasked with reconstructing these features. In the ``with features, w/o feature decoding'' configuration, the encoder processes the graph's attributes, but the feature reconstruction losses ($\mathcal{L}_{nfeat}$, $\mathcal{L}_{cfeat}$) are disabled. Interestingly, we observe a consistent improvement in topological F1 scores across the datasets when side information is provided (e.g., MUTAG improves from $61.29\%$ to $63.61\%$, and AIDS from $84.12\%$ to $86.00\%$). This suggests that chemical or structural attributes act as strong auxiliary signals. They implicitly regularize the topological representation, helping the encoder to construct a more robust geometric embedding even for purely structural tasks.

\paragraph{Impact of Bundle Size.}
We evaluate the impact of bundle size $b \in \{1, 2, 4, 8\}$ on graph reconstruction quality (Figure~\ref{fig:bundle-size-ablation}). Conceptually, increasing $b$ reduces sequence length $T = \lceil n/b \rceil$, which mitigates exposure bias during autoregressive decoding. However, packing multiple node embeddings into a single token creates an information bottleneck. Empirically, the topological F1 score degrades monotonically as $b$ increases, dropping from $84.1\%$ at $b=1$ to $79.5\%$ at $b=8$. In contrast, the graph size prediction error reaches its minimum at $b=2$ ($0.09$ vs.\ $0.23$ at $b=1$). Because the primary objective of \ours{} is high-fidelity topological reconstruction, and an average size error of $0.23$ nodes is practically negligible, we select $b=1$ as our default configuration. Larger bundle sizes ($b > 1$) are reserved as a trade-off mechanism when scaling to very large graphs.

\begin{table}[t]
    \centering
    \begin{tabular}{l|l|c|c}
        Configuration & Dataset & F1 [\%]       & Avg. size diff \\
        \hline
        \multirow{3}{*}{Baseline results}
        & MUTAG   & $61.29 \pm 5.39$ & $0.12 \pm 0.08$    \\
        & AIDS    & $84.12 \pm 1.67$ & $0.23 \pm 0.09$    \\
        & QM9     & $99.38 \pm 0.18$ & $0.00 \pm 0.00$    \\
        \hline
        \multirow{3}{*}{Using VAE}
        & MUTAG   & $60.78 \pm 2.01$ & $0.24 \pm 0.12$    \\
        & AIDS    & $82.17 \pm 1.98$ & $0.61 \pm 0.16$    \\
        & QM9     & $95.88 \pm 4.76$ & $0.00 \pm 0.00$    \\
        \hline
        \multirow{3}{*}{Using only one stacked decoder}
        & MUTAG   & $51.72 \pm 2.91$ & $4.76 \pm 1.69$    \\
        & AIDS    & $72.59 \pm 2.75$ & $4.67 \pm 0.59$    \\
        & QM9     & $98.93 \pm 0.17$ & $0.00 \pm 0.00$    \\
        \hline
        \multirow{3}{*}{With encoded features, without decoding them}
        & MUTAG   & $63.61 \pm 2.85$ & $0.09 \pm 0.09$    \\
        & AIDS    & $86.00 \pm 3.63$ & $0.31 \pm 0.12$    \\
        & QM9     & $98.89 \pm 0.32$ & $0.00 \pm 0.00$    \\
    \end{tabular}
    \caption{The table presents results of ablation study experiments in comparison with the base \ours{} results.}
    \label{tab:ablation_study}
\end{table}

\providecolor{cTfLayersF1}{HTML}{8172B3}
\providecolor{cTfLayersSize}{HTML}{C44E52}

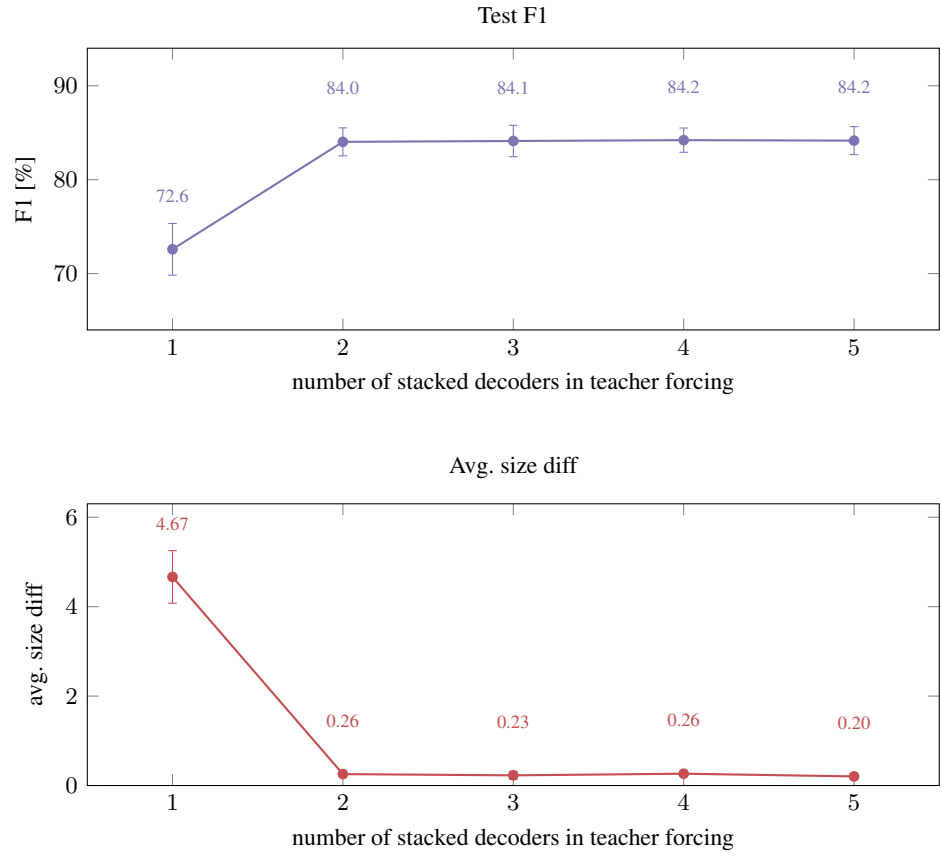
\begin{figure}[t]
    \centering
    \begin{tikzpicture}
        \pgfplotsset{
            every axis/.append style={
                label style={font=\small},
                tick label style={font=\small},
                title style={font=\small},
            }
        }
        \begin{groupplot}[
            group style={group size=1 by 2, vertical sep=2.3cm},
            width=0.92\linewidth,
            height=0.38\linewidth,
            xlabel={number of stacked decoders in teacher forcing},
            xtick={1,2,3,4,5},
            xmin=0.5, xmax=5.5,
            error bars/y dir=both,
            error bars/y explicit,
            every axis plot post/.append style={mark=*, mark size=1.6pt, thick},
            nodes near coords,
            every node near coord/.append style={font=\scriptsize, anchor=south, yshift=14pt},
            point meta=explicit symbolic,
        ]
        \nextgroupplot[
            title={Test F1},
            ylabel={F1~[\%]},
            ymin=64, ymax=94,
        ]
        \addplot+[cTfLayersF1, mark options={fill=cTfLayersF1},
            every node near coord/.append style={text=cTfLayersF1}]
            table[meta=lbl, y error=err] {
            x     y      err    lbl
            1     72.59  2.75   72.6
            2     84.03  1.49   84.0
            3     84.12  1.67   84.1
            4     84.21  1.29   84.2
            5     84.16  1.49   84.2
        };

        \nextgroupplot[
            title={Avg.\ size diff},
            ylabel={avg.\ size diff},
            ymin=0, ymax=6.3,
        ]
        \addplot+[cTfLayersSize, mark options={fill=cTfLayersSize},
            every node near coord/.append style={text=cTfLayersSize}]
            table[meta=lbl, y error=err] {
            x     y       err     lbl
            1     4.6650  0.5876  4.67
            2     0.2556  0.0640  0.26
            3     0.2275  0.0914  0.23
            4     0.2642  0.0417  0.26
            5     0.2040  0.0726  0.20
        };
        \end{groupplot}
    \end{tikzpicture}
    \caption{Effect of the number of stacked decoders in teacher forcing on AIDS test performance: F1 score (left) and average node-count (size) difference (right). Error bars show $\pm$ 1 standard deviation over 5 seeds.}
    \label{fig:tf-ablation}
\end{figure}

\providecolor{cBundleF1}{HTML}{8172B3}
\providecolor{cBundleSize}{HTML}{C44E52}

\begin{figure}[t]
    \centering
    \begin{tikzpicture}
        \pgfplotsset{
            every axis/.append style={
                label style={font=\small},
                tick label style={font=\small},
                title style={font=\small},
            }
        }
        \begin{groupplot}[
            group style={group size=1 by 2, vertical sep=2.3cm},
            width=0.92\linewidth,
            height=0.38\linewidth,
            xlabel={bundle size},
            xmode=log,
            log basis x={2},
            xtick={1,2,4,8},
            xticklabels={1,2,4,8},
            xminorticks=false,
            xmin=0.8, xmax=10,
            error bars/y dir=both,
            error bars/y explicit,
            every axis plot post/.append style={mark=*, mark size=1.6pt, thick},
            nodes near coords,
            every node near coord/.append style={font=\scriptsize, anchor=south, yshift=14pt},
            point meta=explicit symbolic,
        ]
        \nextgroupplot[
            title={Test F1},
            ylabel={F1~[\%]},
            ymin=70, ymax=90,
        ]
        \addplot+[cBundleF1, mark options={fill=cBundleF1},
            every node near coord/.append style={text=cBundleF1}]
            table[meta=lbl, y error=err] {
            x     y      err    lbl
            1     84.10  1.70   84.1
            2     82.92  1.61   82.9
            4     80.36  1.91   80.3
            8     79.46  2.26   79.5
        };

        \nextgroupplot[
            title={Avg.\ size diff},
            ylabel={avg.\ size diff},
            ymin=0, ymax=0.5,
        ]
        \addplot+[cBundleSize, mark options={fill=cBundleSize},
            every node near coord/.append style={text=cBundleSize}]
            table[meta=lbl, y error=err] {
            x     y       err     lbl
            1     0.2300  0.0900  0.23
            2     0.0948  0.0588  0.09
            4     0.1329  0.0813  0.13
            8     0.2879  0.0722  0.29
        };
        \end{groupplot}
    \end{tikzpicture}
    \caption{Effect of bundle size on test performance on AIDS dataset: F1 score (left) and average node-count (size) difference (right). Error bars show $\pm$ 1 standard deviation over 5 seeds.}
    \label{fig:bundle-size-ablation}
\end{figure}
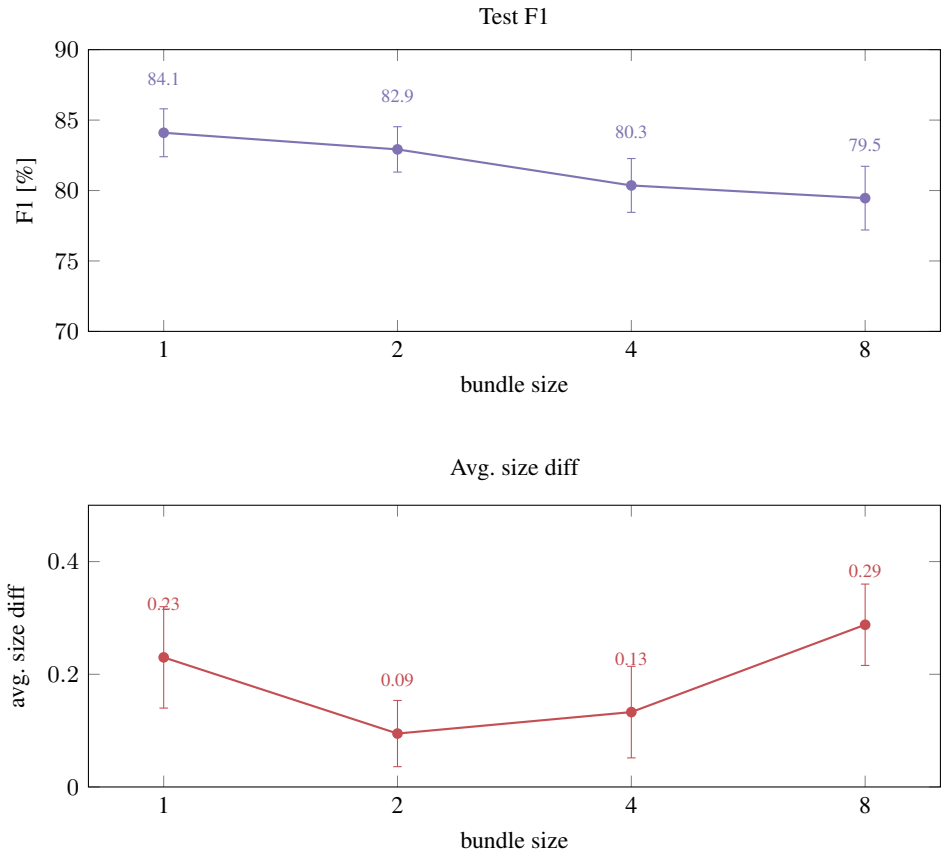

\end{document}